\documentclass[11pt, a4paper, logo]{stockmark}

\usepackage[numbers]{natbib}
\usepackage{fontspec}
\usepackage{xeCJK}

\setCJKsansfont{HaranoAjiGothic-Regular.otf}[BoldFont = HaranoAjiGothic-Bold.otf]
\setCJKmonofont{HaranoAjiGothic-Regular.otf}[BoldFont = HaranoAjiGothic-Bold.otf]

\usepackage{float}    
\usepackage{makecell} 
\usepackage{tocloft}  

\usepackage[hang, flushmargin]{footmisc}
\usepackage{fancyvrb}
\usepackage{framed}
\definecolor{shadecolor}{RGB}{248,248,248}

\newenvironment{Shaded}
  {\begin{snugshade}}
  {\end{snugshade}}

\newenvironment{Highlighting}
  {\VerbatimEnvironment
   \begin{Verbatim}[commandchars=\\\{\},fontsize=\footnotesize]}
  {\end{Verbatim}}

\newcommand{\FunctionTok}[1]{\textcolor[rgb]{0.00,0.44,0.13}{\textbf{#1}}}
\newcommand{\DataTypeTok}[1]{\textcolor[rgb]{0.56,0.13,0.00}{#1}}
\newcommand{\OtherTok}[1]{\textcolor[rgb]{0.00,0.44,0.13}{#1}}
\newcommand{\StringTok}[1]{\textcolor[rgb]{0.25,0.44,0.63}{#1}}
\newcommand{\ErrorTok}[1]{\textcolor[rgb]{0.64,0.00,0.00}{\textbf{#1}}}
\newcommand{\NormalTok}[1]{#1}
\providecommand{\tightlist}{%
  \setlength{\itemsep}{0pt}\setlength{\parskip}{0pt}}

\hypersetup{
    colorlinks=true,
    linkcolor=blue,
    citecolor=blue,
    filecolor=blue,
    urlcolor=blue,
    pdfpagemode=UseNone,
    pdfstartview=FitH
}

\renewcommand{\today}{}

\title{\centering Stockmark-Nemotron-3-Nano-Omni-JapanDocReader: Structured Document Parsing via Capability Injection and Forgetting Control}
\author[*]{
\small
Shi Chen, Hayato Aida, Makoto Morinaga, Shohei Tanaka, Kosuke Arima
\vspace{0.2cm}
\\
\small
\textbf{Stockmark Inc.}
\vspace{0.5cm}
  \\
  {\tiny
  \raggedright{
  \tiny
  \hspace{22.2em}
  \includegraphics[height=1.0em]{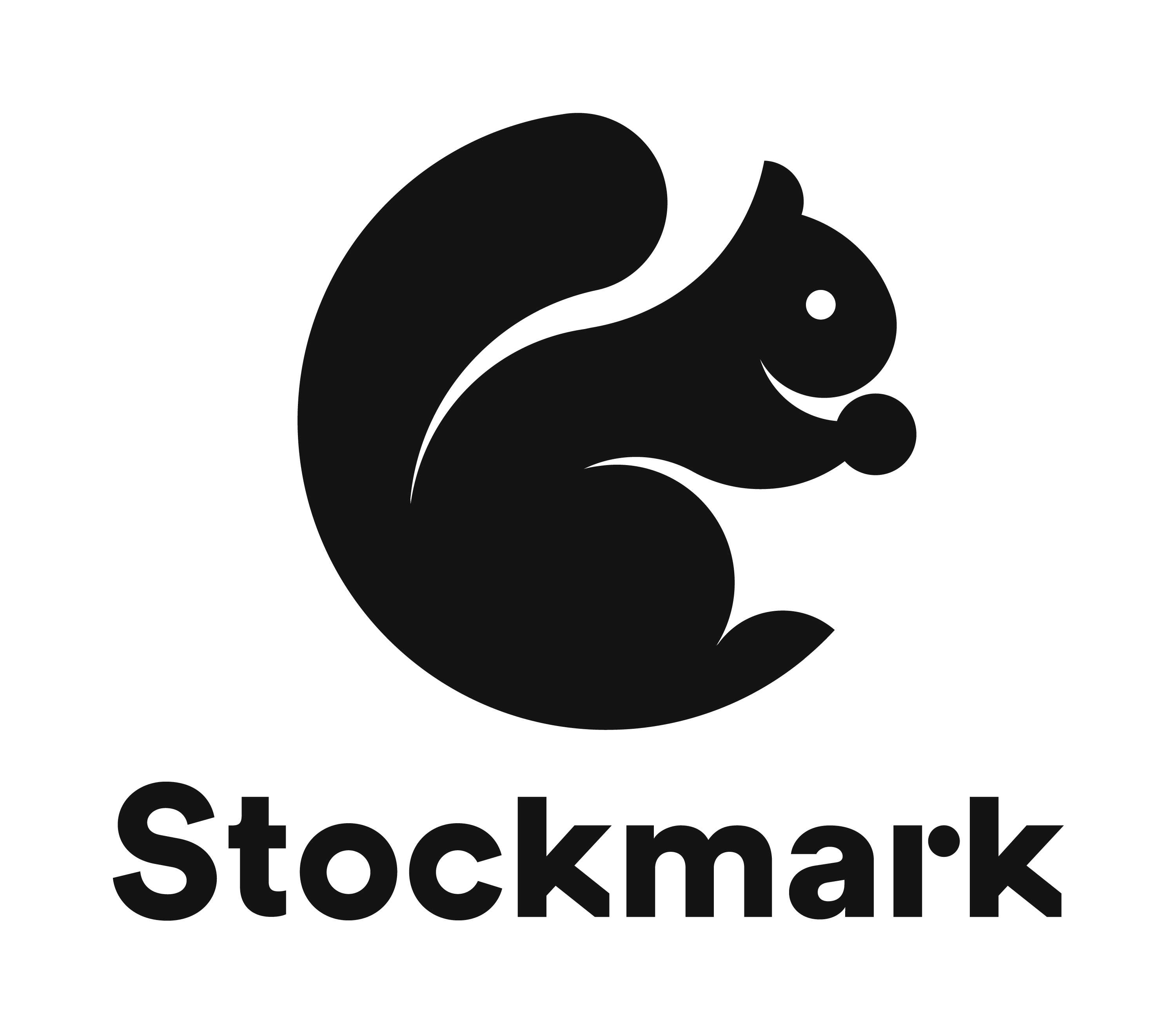} \textbf{Official Website}: \href{https://stockmark.co.jp/}{https://stockmark.co.jp/}  \\
  \tiny
  \hspace{7em}
  \includegraphics[height=1.0em]{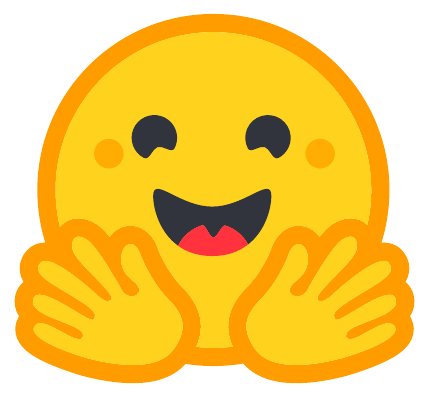} \textbf{Model}: \href{https://huggingface.co/stockmark/Stockmark-Nemotron-3-Nano-Omni-JapanDocReader}{stockmark/Stockmark-Nemotron-3-Nano-Omni-JapanDocReader} \\
  }

  }
}

\begin{abstract}

\vspace{-0.5cm} 

We present \textbf{Stockmark-Nemotron-3-Nano-Omni-JapanDocReader}, a Japanese document understanding model built from Nemotron-3-Nano-Omni-30B-A3B-Reasoning-BF16. The central goal of this work is structured document parsing via capability injection and forgetting control: we inject Japanese structured document parsing capability into a reasoning-oriented multimodal model while preserving its document VQA capability as much as possible. We study parsing-centric SFT, which uses only structured document parsing data; mixed SFT, which combines structured document parsing and VQA data; and parsing-centric RL, which optimizes structured parsing with a task-level reward. Our experiments show that parsing-centric SFT substantially improves structured document parsing performance but causes measurable VQA forgetting. Mixed SFT mitigates this forgetting while preserving nearly the same structured parsing performance. Applying DAPO-based parsing-centric RL on top of the mixed SFT checkpoint further improves structured document parsing beyond the SFT ceiling, producing the final released model. The training data is constructed with a data engine consisting of two complementary synthetic streams: a Japanese Document VQA Stream and a programmatic structured document parsing stream. We also discuss reward design and variance-based prompt filtering for continuous structured document parsing rewards, highlighting their importance for making RL effective in long-reasoning structured document parsing tasks.

\end{abstract}

\begin{document}

\maketitle

\newpage
\setlength{\cftbeforesecskip}{6pt}
\setlength{\cftbeforesubsecskip}{4pt} 
\setcounter{tocdepth}{2}
\tableofcontents

\newpage
\section{Introduction}\label{sec:introduction}

Japanese document understanding requires a model to combine visual perception, OCR-like reading, layout understanding, and multi-step reasoning. A model may need to read dense paragraphs, compare values in charts, interpret tables, recognize formulas, and answer questions that involve aggregation, conditional judgment, or arithmetic. Reasoning-oriented vision-language models (VLMs) are naturally attractive for document VQA because they can allocate a long reasoning trace before producing a final answer. However, structured document parsing imposes a different requirement: the model must convert the entire page into a structured representation.

In this work, structured document parsing refers to the task of converting a document image into a JSON object with the following schema:

\begin{Shaded}
\begin{Highlighting}
\FunctionTok{\{}
  \DataTypeTok{"document\_structure"}\FunctionTok{:} \OtherTok{[}
    \FunctionTok{\{}
      \DataTypeTok{"class"}\FunctionTok{:} \StringTok{"title | heading | text | table | list | picture | formula"}\FunctionTok{,}
      \DataTypeTok{"bbox"}\FunctionTok{:} \OtherTok{[}\ErrorTok{x1}\OtherTok{,} \ErrorTok{y1}\OtherTok{,} \ErrorTok{x2}\OtherTok{,} \ErrorTok{y2}\OtherTok{]}\FunctionTok{,}
      \DataTypeTok{"contents"}\FunctionTok{:} \StringTok{"..."}\FunctionTok{,}
      \DataTypeTok{"caption"}\FunctionTok{:} \StringTok{"..."}
    \FunctionTok{\}}
  \OtherTok{]}
\FunctionTok{\}}
\end{Highlighting}
\end{Shaded}

The seven supported layout classes are \texttt{title}, \texttt{heading}, \texttt{text}, \texttt{list}, \texttt{table}, \texttt{picture}, and \texttt{formula}. For tables, the \texttt{contents} field is represented as HTML. For formulas, the \texttt{contents} field is represented as LaTeX. For pictures, the \texttt{contents} field is a textual description of the visual content, and \texttt{caption} is optional. Bounding boxes are represented in a normalized coordinate convention in the training data. This task differs from Japanese document VQA in both supervision and output structure. Japanese Document VQA depends strongly on the model's original reasoning distribution, answer style, and ability to perform multi-step analysis. Structured document parsing, in contrast, asks the model to emit a complete page-level representation, where each element must have the correct class, approximate location, and content. This difference creates a constrained post-training problem:

\begin{quote}
Can we inject Japanese structured document parsing capability into a reasoning-oriented multimodal model while preserving its document VQA capability as much as possible?
\end{quote}

We formulate this problem as \textbf{structured document parsing via capability injection and forgetting control}. The goal is not only to teach the model a new output format, but also to preserve the base model's existing Japanese document VQA behavior. This is challenging because structured parsing and VQA impose different supervision signals. Parsing requires dense page-level outputs with element classes, localization, reading order, and structured contents, whereas VQA requires question-conditioned reasoning and concise answers. Optimizing strongly for one behavior can easily degrade the other.

We start from Nemotron-3-Nano-Omni-30B-A3B-Reasoning-BF16~\cite{deshmukh2026nemotron}, a Mamba2-Transformer hybrid MoE multimodal reasoning model with approximately 31B total parameters and roughly 3B active parameters per token. The base model supports video, audio, image, and text inputs and produces text outputs. It also supports long-context inputs and uses thinking mode by default. In this report, \textbf{thinking mode} refers to a response format in which the model produces an English \texttt{<think>} reasoning trace before the final answer or the final \texttt{document\_structure} JSON. By contrast, \textbf{instruct mode} refers to direct-answer training and decoding without an explicit \texttt{<think>} trace; for structured document parsing, the model is trained to output the JSON directly. The comparison between thinking mode and instruct mode therefore changes the presence of reasoning traces and the corresponding response format.

A key contribution of this work is a data engine for Japanese document post-training. The data engine contains two synthetic data streams: a Japanese Document VQA Stream for preserving document VQA behavior, and a programmatic structured document parsing stream for injecting page-level parsing capability. The engine is designed not only to increase data volume, but also to control supervision quality, document diversity, reasoning format, and the balance between capability injection and forgetting. We describe the data engine in detail in Section~\ref{sec:data-engine}. With the data engine, we study a two-stage post-training recipe consisting of supervised fine-tuning (SFT) and reinforcement learning (RL). In the SFT stage, we compare two strategies: parsing-centric SFT, which uses only structured document parsing data and directly injects parsing capability, and mixed SFT, which combines parsing and VQA data to mitigate forgetting while maintaining parsing performance. In the RL stage, we apply parsing-centric RL to further optimize structured parsing performance using a matched, application-aware reward and variance-based prompt filtering. The final model, Stockmark-Nemotron-3-Nano-Omni-JapanDocReader, is obtained by mixed SFT followed by DAPO-based parsing-centric RL and is released on Hugging Face. Our experiments show that parsing-centric SFT effectively injects structured parsing capability but causes measurable VQA forgetting. Mixed SFT substantially mitigates this forgetting while maintaining almost the same parsing performance. DAPO-based parsing-centric RL further improves structured document parsing beyond the mixed SFT ceiling, producing the best model in this study, at the cost of additional VQA drift.

In summary, this report makes three main contributions. First, we build a data engine that supports both Japanese Document VQA and structured parsing data with ground truth obtained by construction. Second, we analyze the capability-injection and forgetting-control trade-off through parsing-centric SFT and mixed SFT. Third, we show that DAPO-based parsing-centric RL can push structured document parsing performance beyond the SFT ceiling when combined with matched rewards and variance-based prompt filtering.

\section{Data Engine}\label{sec:data-engine}

This section describes the data engine used to construct the training data for this work. The engine consists of two largely independent synthetic data streams:

\begin{enumerate}
\def\labelenumi{\arabic{enumi}.}
\tightlist
\item
a Japanese Document VQA Stream, and
\item
a programmatic structured document parsing stream.
\end{enumerate}

The two streams are intentionally decoupled because they serve different roles in post-training. The VQA stream produces examples for reading and reasoning over Japanese document images, and is used to preserve the base model's document VQA capability. The structured document parsing stream produces examples for extracting a complete \texttt{document\_structure} JSON object from a page image, and is used to inject structured parsing capability.

\begin{figure}[H]
\centering
\includegraphics[width=0.95\linewidth]{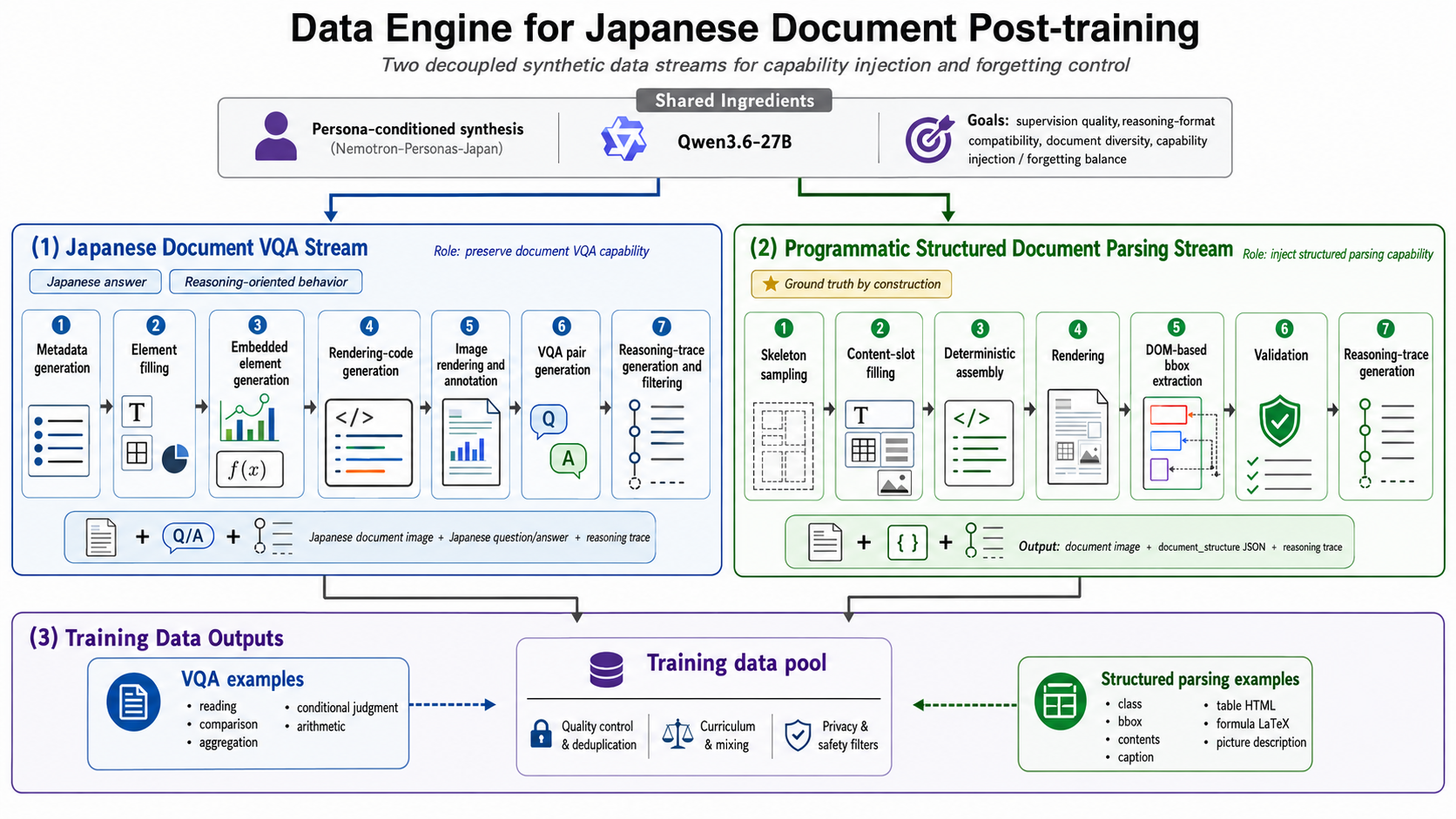} 
\caption{
    \centering
     Overview of the Data Engine pipeline.
}
\label{fig:data_engine}
\end{figure}

In both streams, we use persona-conditioned synthesis based on Nemotron-Personas-Japan dataset~\cite{nvidia_nemotron_personas_japan_2025} to diversify topics, occupations, industries, document styles, and professional contexts. The main LLM used inside the data engine is Qwen3.6-27B~\cite{qwen36}. Depending on the stage, Qwen3.6-27B is used to generate metadata, textual content, questions, answers, content slots, and English reasoning traces. The overall design goal of the data engine is not only to increase data volume, but also to control supervision quality, reasoning format, document diversity, and the balance between capability injection and forgetting. The overall pipeline is illustrated in Figure~\ref{fig:data_engine}.

\subsection{Japanese Document VQA Stream}\label{sec:english-reasoning-japanese-vqa-pipeline}

The VQA stream follows an \textbf{English reasoning, Japanese answer} design, which is motivated by the behavior of the base model. Nemotron-3-Nano-Omni-30B-A3B-Reasoning is a reasoning-oriented model that naturally produces English reasoning traces. Forcing it to reason in Japanese may shift the model away from its native reasoning distribution. Therefore, the stream keeps the document image, question, and final answer in Japanese, while using English reasoning traces. This preserves a reasoning style closer to the base model while keeping the final answer compatible with Japanese VQA output.

The VQA stream is a multi-stage synthesis process. It first constructs document images from structured metadata, then generates VQA pairs and reasoning traces over those images.

\begin{itemize}
\item
\textbf{Metadata generation.} Qwen3.6-27B generates a structured metadata skeleton for each sample, including the document category, document type, section structure, chart or table specifications, and other high-level attributes. Persona conditioning from Nemotron-Personas-Japan dataset is used to diversify domains, roles, industries, and writing styles.
\item
\textbf{Element filling.} Qwen3.6-27B fills the metadata skeleton with concrete Japanese content. This includes headings, body text, chart labels, table cells, numerical values, domain-specific terminology, and other document contents. Separating metadata generation from element filling makes the page structure easier to control while retaining content diversity.
\item
\textbf{Embedded element generation.} When a document contains embedded charts, tables, diagrams, or subfigures, Qwen3.6-27B generates the corresponding internal contents. This stage is important for VQA because many questions require reading small embedded visual elements or combining information across text and figures.
\item
\textbf{Rendering-code generation.} The filled document specification is converted into executable rendering code or markup. Qwen3.6-27B writes the rendering code, while deterministic checks validate that the output is executable and compatible with the target renderer.
\item
\textbf{Image rendering and annotation.} The renderer is executed to produce the final document image. The rendering process also provides element-level metadata and bounding boxes, which are used by downstream trajectory planning and quality control.
\item
\textbf{VQA pair generation.} Qwen3.6-27B generates Japanese question-answer pairs over the rendered document image and its metadata. The questions cover direct reading, comparison, aggregation, conditional judgment, and simple arithmetic, with an emphasis on tasks that benefit from explicit reasoning.
\item
\textbf{Reasoning-trace generation and filtering.} Qwen3.6-27B generates reasoning traces for the questions. The generated traces are filtered using programmatic checks and judge models to remove invalid answers, degenerate reasoning, language mismatches, and low-quality trajectories.
\end{itemize}

\subsection{Programmatic Structured Document Parsing Stream}\label{sec:programmatic-structured-parsing-pipeline}

The structured document parsing stream is designed around ground truth by construction, which first samples a programmatic layout skeleton and then asks the LLM to fill textual or semantic content slots. The page is assembled and rendered deterministically, so the JSON annotation is tied directly to the rendering source.

The stream consists of the following stages.

\begin{itemize}
\item
\textbf{Skeleton sampling.} The pipeline samples a programmatic page plan containing the layout family, target element density, text chunks, tables, figures, formulas, repeated structures, and style parameters. This stage controls difficult supervision factors such as dense layouts, multi-column pages, vertical Japanese writing, long forms, and complex tables.
\item
\textbf{Content-slot filling.} Qwen3.6-27B fills the slots defined by the skeleton with Japanese content, including titles, headings, body text, list items, table labels, figure descriptions, and semantic contents. The LLM is responsible for producing natural and diverse content, while the skeleton controls the structure.
\item
\textbf{Deterministic assembly.} The page is assembled as HTML or renderer-specific markup. Each ground-truth element is marked with a \texttt{data-gt} identifier so that the rendered element and the JSON target are tied to the same source object.
\item
\textbf{Rendering.} The assembled page is rendered with a browser-based or renderer-specific backend to produce the final document image.
\item
\textbf{DOM-based bbox extraction.} Bounding boxes are read directly from the rendered DOM or rendering object. This avoids post-hoc visual extraction noise and ensures that the JSON annotation is consistent with the rendered page.
\item
\textbf{Validation.} The pipeline checks JSON schema validity, class labels, bounding-box ranges, table parseability, formula format, and content sanity. Invalid pages are discarded before training data construction.
\item
\textbf{Reasoning-trace generation.} Structured document parsing examples also contain English reasoning traces. These traces are generated by Qwen3.6-27B using few-shot prompts. The few-shot examples demonstrate how to inspect the page, enumerate layout elements, reason about element classes, approximate locations, reading order, table structures, formulas, and pictures, and then emit the final \texttt{document\_structure} JSON. This is different from simply inserting a fixed format string into \texttt{\textless{}think\textgreater{}}: the reasoning trace is generated as a page-specific explanation of the parsing target.
\end{itemize}

The central advantage of this stream is that difficult supervision signals are controlled by code rather than by post-hoc extraction. Element density can be targeted explicitly. Tables are generated from the same HTML tree that is rendered into the image, so table supervision remains consistent with the visual page. Formula LaTeX can be validated through rendering. Bounding boxes are obtained from the rendered layout itself. The stream also supports difficult document families such as dense multi-column pages, vertical Japanese writing, newspaper-style layouts, handwritten-style pages, long forms, complex tables with row/column spans, and formula-heavy pages.

\section{Training Strategy}\label{sec:training-recipe-datasets}

\begin{figure}[H]
\centering
\includegraphics[width=0.95\linewidth]{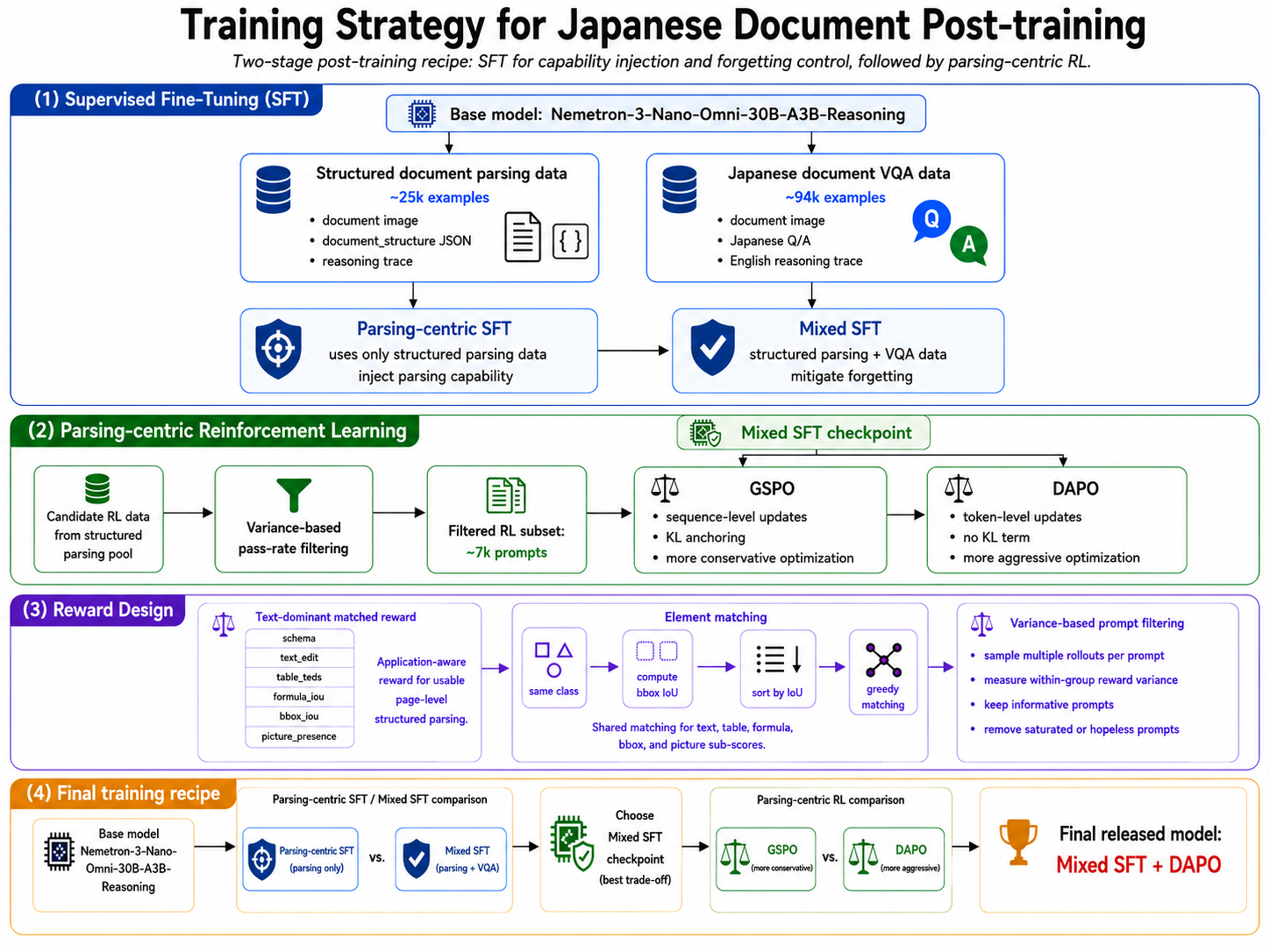} 
\caption{
    \centering
     Overview of the training strategy.
}
\label{fig:training_strategy}
\end{figure}

Figure~\ref{fig:training_strategy} provides an overview of the training strategy used in this work. The overall recipe consists of two stages: supervised fine-tuning for capability injection and forgetting control, followed by parsing-centric RL for further improving structured document parsing performance.

\subsection{Supervised Fine-Tuning}\label{sec:supervised-fine-tuning}

The SFT stage is designed to study the trade-off between injecting structured document parsing capability and preserving the base model's document VQA capability. We therefore compare two SFT strategies: parsing-centric SFT and mixed SFT. Parsing-centric SFT uses only structured document parsing examples. This setting is intended to separate the parsing objective from VQA and to directly supervise the model to output document elements, including class labels, bounding boxes, reading order, text contents, tables, formulas, and picture descriptions. In contrast, mixed SFT combines structured document parsing data with Japanese document VQA data. The goal of Mixed SFT is to inject structured parsing capability while reducing forgetting of the original VQA behavior.

\textbf{Training data.} The VQA data pool used in this work contains approximately 94k examples. Each example consists of a Japanese document image, a Japanese question-answer pair with English reasoning trace. The questions emphasize multi-step reasoning, including comparison, aggregation, conditional judgment, and simple calculations. The structured document parsing data pool contains approximately 25k examples. Each example consists of a document image, a structured JSON annotation in the \texttt{document\_structure} schema, and an English reasoning trace. The annotations include the seven layout classes \texttt{title}, \texttt{heading}, \texttt{text}, \texttt{list}, \texttt{table}, \texttt{picture}, and \texttt{formula}. Tables are represented as HTML, formulas as LaTeX, and pictures include textual descriptions. The data intentionally includes difficult layouts such as dense pages, vertical writing, handwritten-style pages, newspaper-style layouts, complex tables, and formula-heavy pages.

\textbf{Training configuration.} All SFT experiments in this report are conducted using Megatron-Bridge\footnote{https://github.com/NVIDIA-NeMo/Megatron-Bridge} with the same full-parameter SFT hyperparameters as shown in Table~\ref{tab:sft-setting}.

\begin{table}[h]
\centering
 \fontsize{8}{8}\selectfont
\renewcommand{\arraystretch}{1.2}
\begin{tabular}{l|c}
\toprule
\textbf{Settings} & \textbf{SFT}  \\  \midrule
Trainable modules & language model + vision encoder + vision projector  \\
Context length & 24576 \\
Global batch size & 64 \\
Micro batch size & 1 \\
Learning rate & 6e-6 \\
Parallelism & TP=2, EP=8, ETP=1, PP=1, CP=1 \\
Epoch & 1  \\ \bottomrule
\end{tabular}
\caption{Training settings for SFT stage.}
\label{tab:sft-setting}
\end{table}

\subsection{Reinforcement Learning}\label{sec:structured-parsing-reinforcement-learning}

After SFT, we apply parsing-centric RL using structured document parsing prompts only. The term parsing-centric RL emphasizes that the RL stage is optimized for the parsing task, rather than for VQA-style question answering. We compare GSPO~\cite{zheng2025group} and DAPO~\cite{yu2026dapo} for parsing-centric RL. One relevant difference lies in the granularity at which policy updates are defined.

GSPO performs optimization based on a sequence-level importance ratio. For a prompt \(x\) and a generated response \(y_i=(y_{i,1},\ldots,y_{i,T_i})\), the sequence-level ratio is computed from the average log-probability ratio over the whole response:

\[
\rho_i^{\mathrm{seq}}(\theta)
=
\exp\left(
\frac{1}{T_i}
\sum_{t=1}^{T_i}
\log
\frac{
\pi_{\theta}(y_{i,t}\mid x,y_{i,<t})
}{
\pi_{\mathrm{old}}(y_{i,t}\mid x,y_{i,<t})
}
\right).
\]

Given \(G\) rollouts for the same prompt, we compute a group-normalized advantage \(\hat{A}_i\) from the structured parsing reward. The GSPO objective can be written as

\[
\mathcal{L}_{\mathrm{GSPO}}(\theta)
=
-\mathbb{E}
\left[
\frac{1}{G}
\sum_{i=1}^{G}
\min\left(
\rho_i^{\mathrm{seq}}(\theta)\hat{A}_i,\,
\mathrm{clip}\left(
\rho_i^{\mathrm{seq}}(\theta),
1-\epsilon,
1+\epsilon
\right)\hat{A}_i
\right)
\right]
+
\beta\,
D_{\mathrm{KL}}
\left(
\pi_{\theta}
\,\|\, 
\pi_{\mathrm{ref}}
\right).
\]

Here, \(\pi_{\mathrm{old}}\) is the rollout policy, and \(\pi_{\mathrm{ref}}\) is the reference policy, which is the mixed SFT checkpoint in our setting. Because GSPO evaluates policy change at the level of an entire generated sequence, it can be advantageous for stabilizing training in MoE models, where token-level ratio variation can become large. In our setup, GSPO also uses KL anchoring to keep the policy close to the mixed SFT checkpoint.

In contrast, DAPO uses a token-level policy-gradient loss. The token-level importance ratio is defined as

\[
\rho_{i,t}^{\mathrm{tok}}(\theta)
=
\frac{
\pi_{\theta}(y_{i,t}\mid x,y_{i,<t})
}{
\pi_{\mathrm{old}}(y_{i,t}\mid x,y_{i,<t})
}.
\]

The DAPO objective used in this work can be written as

\[
\mathcal{L}_{\mathrm{DAPO}}(\theta)
=
-\mathbb{E}
\left[
\frac{1}{\sum_{i=1}^{G} T_i}
\sum_{i=1}^{G}
\sum_{t=1}^{T_i}
\min\left(
\rho_{i,t}^{\mathrm{tok}}(\theta)\hat{A}_i,\,
\mathrm{clip}\left(
\rho_{i,t}^{\mathrm{tok}}(\theta),
1-\epsilon_{\mathrm{low}},
1+\epsilon_{\mathrm{high}}
\right)\hat{A}_i
\right)
\right].
\]

Unlike GSPO, DAPO does not use a KL term in our experiments. Instead, it controls the update magnitude mainly through the learning rate, PPO clipping, clip-higher, reward scaling, and reward shaping. With these configurations, the GSPO variant is more strongly regularized toward the mixed-SFT checkpoint, whereas the DAPO variant is allowed to move farther from the SFT checkpoint in order to improve structured document parsing performance.

\textbf{Training data.} The candidate RL data is derived from the structured document parsing pool and then filtered using the variance-based pass-rate filtering procedure described below. The filtered RL subset contains approximately 7k structured document parsing prompts. This filtering step was used before the parsing-centric RL experiments because the reward is continuous and not every prompt provides useful group-relative learning signal.

\textbf{Training configuration.} All RL experiments in this report are conducted using NeMo-RL\footnote{https://github.com/NVIDIA-NeMo/RL} with the same full-parameter RL hyperparameters as shown in Table~\ref{tab:rl-setting}

\begin{table}[h]
\centering
\fontsize{8}{8}\selectfont
\renewcommand{\arraystretch}{1.15}
\setlength{\tabcolsep}{4pt}
\begin{tabular}{l|c}
\toprule
\textbf{Settings} & \textbf{RL} \\
\midrule
Trainable modules & language model \\
Context length & 24576 \\
Global batch size & 256 = 32 prompts $\times$ 8 generations \\
Micro batch size & 1 \\
Learning rate & 3e-6 \\
Sampling temperature & 0.8 \\
Max new tokens & 20480 \\
Reasoning token budget & 16384 \\
\bottomrule
\end{tabular}
\caption{Training settings for RL stage.}
\label{tab:rl-setting}
\end{table}

\subsubsection{Reward Design}\label{sec:reward-design-evaluation-alignment-and-application-alignment}

The purpose of parsing-centric RL is to improve page-level structured document parsing quality beyond the SFT ceiling. Token-level cross-entropy does not directly correspond to the quality of a parsed document structure. For example, a single extra or missing element near the beginning of the \texttt{document\_structure} array can shift all later elements under index-wise comparison, even if many of the model's extracted elements are otherwise reasonable. Similarly, a response may contain mostly correct text, tables, or formulas, but still be unusable in practice if it violates the JSON schema, localizes elements poorly, misses important regions, or omits descriptions for pictures.

We therefore design a representation-aware reward that first enforces a strict validity gate and then computes task-aligned similarity on a canonicalized document-structure representation. Let \(y\) be the model output and \(y^*\) be the reference annotation. The canonicalization function \(\phi(\cdot)\) parses a valid output into a set of document elements, where each element contains a class label, bounding box, contents, and optional caption. We denote the canonicalized prediction and reference as

\[
P = \phi(y), \qquad G = \phi(y^*).
\]

The final reward is defined as

\[
R(y,y^*)
=
\mathrm{Valid}_{\mathrm{doc}}(y)
\cdot
\sum_{k \in \mathcal{K}}
w_k S_k(P,G),
\]

where \(\mathcal{K}\) is the set of reward components, \(S_k(P,G)\in[0,1]\) is the normalized sub-score for component \(k\), and \(w_k\) is its weight. The weights satisfy \(\sum_{k\in\mathcal{K}} w_k = 1\). The concrete components and weights used in our experiments are summarized in Table~\ref{tab:reward-design-components}.

Here, \(\mathrm{Valid}_{\mathrm{doc}}(y)\in\{0,1\}\) is a strict validity gate. It is set to zero if the output cannot be parsed as JSON, does not contain a top-level \texttt{document\_structure} array, is truncated, or exhibits severe degeneration. When this gate is zero, the total reward is zero. If the top-level structure is valid, all remaining terms are computed on the canonicalized representation. This design makes the reward both strict and dense: unusable outputs receive zero reward, while valid structured outputs receive continuous feedback from schema, content, localization, and picture-description components.

\begin{table}[h]
\centering
\fontsize{8}{8}\selectfont
\renewcommand{\arraystretch}{1.2}
\setlength{\tabcolsep}{4pt}
\begin{tabular}{l|c|p{7.6cm}}
\toprule
\textbf{Component} & \textbf{Weight} & \textbf{Definition} \\
\midrule
\(\mathrm{Valid}_{\mathrm{doc}}\) &
-- &
Strict validity gate for usable structured output. The output must be parseable as JSON and must contain a top-level \texttt{document\_structure} array. Invalid JSON, missing \texttt{document\_structure}, truncation, or severe degeneration receives zero total reward. \\

\(S_{\mathrm{schema}}\) &
0.15 &
Element-level schema validity after the top-level validity gate. It checks whether emitted elements use valid classes and well-formed fields such as \texttt{bbox}, \texttt{contents}, and optional \texttt{caption}. Malformed elements are excluded from matching. \\

\(S_{\mathrm{text}}\) &
0.45 &
Mean text similarity over all ground-truth text-class elements, including \texttt{title}, \texttt{heading}, \texttt{text}, and \texttt{list}. Unmatched ground-truth elements receive zero. \\

\(S_{\mathrm{table}}\) &
0.20 &
Mean TEDS over all ground-truth table elements after converting table contents to HTML. Unmatched ground-truth tables receive zero. \\

\(S_{\mathrm{formula}}\) &
0.05 &
Mean formula similarity over all ground-truth formula elements, using CDM when available and LaTeX edit similarity as fallback. Unmatched ground-truth formulas receive zero. \\

\(S_{\mathrm{bbox}}\) &
0.10 &
Mean IoU over all valid ground-truth layout elements. Unmatched ground-truth elements receive zero. \\

\(S_{\mathrm{picture}}\) &
0.05 &
Fraction of ground-truth picture elements that are matched to predicted picture elements with non-empty \texttt{contents}. \\
\bottomrule
\end{tabular}
\caption{Components and weights of the representation-aware reward for structured document parsing.}
\label{tab:reward-design-components}
\end{table}

This design makes the reward both strict and dense. The \(\mathrm{Valid}_{\mathrm{doc}}\) gate prevents unusable outputs from receiving partial credit, while the matched similarity terms provide continuous feedback for valid outputs. The reward is intentionally broader than simple content matching: it evaluates schema validity, element-level content, localization, and picture-description availability, all of which are required for usable page-level structured parsing.

\subsubsection{Element Matching}\label{sec:element-matching}

A central part of the reward is \textbf{element matching}. All content and localization sub-scores share the same predicted-to-ground-truth element matching. This is designed to remove artificial order noise and to prevent proxy divergence between the training reward and the evaluation protocol.

Given ground-truth elements \(G=\{g_i\}\) and predicted elements \(P=\{p_j\}\), the matching procedure is:

\begin{enumerate}
\def\labelenumi{\arabic{enumi}.}
\tightlist
\item
  Discard predicted and ground-truth elements without valid bounding boxes.
\item
  Construct candidate pairs \((g_i,p_j)\) only when the two elements have the same class.
\item
  Compute the bbox IoU for every candidate pair.
\item
  Keep candidate pairs whose IoU is at least \(\tau_{\mathrm{IoU}}\).
\item
  Sort candidate pairs by descending IoU.
\item
  Greedily assign pairs, ensuring that each ground-truth element and each predicted element is used at most once.
\end{enumerate}

Formally, the resulting matching is a partial map

\[
M: i \mapsto j,
\]

where \(g_i\) and \(p_j\) have the same class and no ground-truth or predicted element is matched more than once. The IoU threshold is intentionally lenient. A loose match still receives a low bbox score and may also receive a low content score, but the match prevents a single insertion or deletion from misaligning all subsequent elements.

Importantly, the denominators of the content and localization sub-scores are defined over ground-truth elements, not only over matched elements. Therefore, unmatched ground-truth elements explicitly reduce the reward.

For text-class elements, let \(G_{\mathrm{text}}\) be the set of ground-truth elements whose class is one of \(\{\texttt{title}, \texttt{heading}, \texttt{text}, \texttt{list}\}\). The text sub-score is defined as

\[
S_{\mathrm{text}}(P,G)
=
\frac{1}{|G_{\mathrm{text}}|}
\sum_{g_i \in G_{\mathrm{text}}}
\mathbb{1}[i \in \mathrm{dom}(M)]
\cdot
\mathrm{sim}_{\mathrm{text}}(g_i,p_{M(i)}),
\]

where \(\mathrm{sim}_{\mathrm{text}} = 1-\mathrm{NED}\), clipped to \([0,1]\). Thus, an unmatched ground-truth text element contributes zero.

For tables, let \(G_{\mathrm{table}}\) be the set of ground-truth table elements. The table sub-score is

\[
S_{\mathrm{table}}(P,G)
=
\frac{1}{|G_{\mathrm{table}}|}
\sum_{g_i \in G_{\mathrm{table}}}
\mathbb{1}[i \in \mathrm{dom}(M)]
\cdot
\mathrm{TEDS}(g_i,p_{M(i)}),
\]

where table contents are converted to HTML before computing TEDS. The TEDS score is normalized to \([0,1]\), and unmatched ground-truth tables contribute zero.

For formulas, let \(G_{\mathrm{formula}}\) be the set of ground-truth formula elements. The formula sub-score is

\[
S_{\mathrm{formula}}(P,G)
=
\frac{1}{|G_{\mathrm{formula}}|}
\sum_{g_i \in G_{\mathrm{formula}}}
\mathbb{1}[i \in \mathrm{dom}(M)]
\cdot
\mathrm{sim}_{\mathrm{formula}}(g_i,p_{M(i)}),
\]

where \(\mathrm{sim}_{\mathrm{formula}}\) uses CDM when available and LaTeX edit similarity as fallback. The score is normalized to \([0,1]\), and unmatched ground-truth formulas contribute zero.

For localization, let \(G_{\mathrm{valid}}\) be the set of ground-truth elements with valid bounding boxes. The bbox sub-score is

\[
S_{\mathrm{bbox}}(P,G)
=
\frac{1}{|G_{\mathrm{valid}}|}
\sum_{g_i \in G_{\mathrm{valid}}}
\mathbb{1}[i \in \mathrm{dom}(M)]
\cdot
\mathrm{IoU}(g_i,p_{M(i)}).
\]

Thus, missed ground-truth elements reduce the localization reward even when other elements are matched correctly.

For picture elements, let \(G_{\mathrm{picture}}\) be the set of ground-truth picture elements. The picture-presence sub-score is

\[
S_{\mathrm{picture}}(P,G)
=
\frac{1}{|G_{\mathrm{picture}}|}
\sum_{g_i \in G_{\mathrm{picture}}}
\mathbb{1}
\left[
i \in \mathrm{dom}(M)
\land
\mathrm{nonempty}\left(\mathrm{contents}(p_{M(i)})\right)
\right].
\]

This score measures the fraction of ground-truth pictures that are matched to predicted picture elements with non-empty descriptions.

When a class-specific ground-truth set is empty, the corresponding class-specific score is treated as satisfied if the prediction also contains no valid element of that class, and as zero if the prediction contains spurious elements of that class. For \(S_{\mathrm{text}}\), this convention is applied to the union of \texttt{title}, \texttt{heading}, \texttt{text}, and \texttt{list}. For \(S_{\mathrm{bbox}}\), the same convention is used in the rare case where no valid ground-truth layout element exists.

Unmatched predicted elements are false positives. In the current reward, false positives are not assigned a separate precision penalty when the corresponding ground-truth class is non-empty; the reward is primarily recall-oriented because the denominators are defined over ground-truth elements. False positives can still affect the reward indirectly through greedy matching competition, and they are penalized in the class-absent case described above. We leave an explicit precision-oriented false-positive penalty as a possible future refinement.

\subsubsection{Variance-based Pass-rate Filtering for Continuous Structured Rewards}\label{sec:variance-based-pass-rate-filtering-for-continuous-structured-rewards}

Before parsing-centric RL, we filter prompts using the current SFT initialization policy. This procedure is analogous to pass-rate filtering in VQA, but structured document parsing reward is continuous and multi-dimensional, so a binary pass/fail decision is not appropriate. Instead, the filter targets the quantity that directly drives group-relative RL: \textbf{within-group reward variance}.

For each prompt \(i\), we sample \(G\) rollouts from the SFT policy and score each rollout with the same structured document parsing reward used during training. We then compute:

\[
\mu_i = \frac{1}{G}\sum_{j=1}^{G} r_{ij},
\qquad
\sigma_i = \sqrt{\frac{1}{G}\sum_{j=1}^{G}(r_{ij}-\mu_i)^2}.
\]

A prompt is retained when:

\[
\sigma_i > \sigma_{\mathrm{floor}}
\quad \text{and} \quad
\mu_{\min} < \mu_i < \mu_{\max}.
\]

The intuition is simple. If all rollouts for a prompt receive almost the same reward, the group-relative advantage is nearly zero, so the prompt provides little gradient signal. If \(\mu_i\) is too high, the prompt is already solved and leaves little room for RL. If \(\mu_i\) is too low, the prompt may be unlearnable or may contain noisy ground truth. The retained prompts are therefore those that are neither hopeless nor saturated, and that expose meaningful quality differences among rollouts.

The filtering script also profiles text-edit reward variance, schema validity rate, and schema-valid-only variance. This helps distinguish true document parsing skill variation from variance caused only by JSON validity flips. After element matching removes order-induced noise, the observed reward standard deviation becomes lower but more meaningful. For this reason, the standard-deviation floor should be calibrated from the observed distribution, for example using a percentile-based threshold, rather than reusing an old absolute value.

\section{Experiments}\label{sec:experiments}

\subsection{Evaluation Setup}\label{sec:evaluation-setup}

\subsubsection{Structured Document Parsing Evaluation}\label{sec:structured-parsing-evaluation}

We evaluate structured document parsing performance on OmniDocBench-JASyn~\cite{stockmark_omnidocbenchjasyn_2026}, which follows the OmniDocBench~\cite{ouyang2025omnidocbench} format, a Japanese document parsing benchmark containing 520 synthetic Japanese document images covering 14 document types. The benchmark includes visual degradations such as skew, print noise, and artifacts to approximate scanned documents.

We report the following summary metric:

\[
\mathrm{DocParse\text{-}Overall}
=
\frac{(1-\mathrm{Text\ Edit})\times 100 + \mathrm{Table\ TEDS} + \mathrm{Formula\ CDM}}{3}
\]

This metric summarizes three core components of document parsing quality: pure text extraction, table structure extraction, and formula recognition. \texttt{Text\ Edit} is the normalized edit distance~\cite{levenshtein1966binary} for pure text, where lower is better. We compute normalized edit distance at the sample level and then average the scores across samples. For tables, all predicted and ground-truth tables are converted into HTML format before evaluation. Table similarity is then measured using Tree-Edit-Distance-based Similarity~\cite{zhong2020image} (\texttt{Table\ TEDS}), where higher values indicate more accurate table structure reconstruction. For formulas, we use the Character Detection Matching metric~\cite{wang2024cdm} (\texttt{Formula\ CDM}), which evaluates formula recognition quality at the character level; higher values are better.

In addition to DocParse-Overall, we also report \texttt{Reading Order Edit}. Reading order is evaluated using normalized edit distance, where lower is better. This metric measures whether the model outputs document elements in a coherent order, which is important for downstream use.

\subsubsection{Japanese document VQA evaluation}\label{sec:japanese-document-vqa-evaluation}

We evaluate Japanese document VQA performance as the arithmetic mean of the following benchmarks.

\begin{itemize}
\tightlist
\item
\textbf{JA-Business-Doc-RQ-Bench}~\cite{stockmark_jabusinessdocrqbench_2026}, a benchmark dataset for evaluating multi-step reasoning ability on visually rich Japanese business documents. The dataset contains synthetic but realistic business-related visual documents paired with manually written VQA tasks. All images are synthetically generated, whereas the question--answer pairs are manually annotated. The images often contain dense information and require multiple reasoning steps to answer the questions correctly.
\item
\textbf{JGraphQA-Refined}, the refined version of JGraphQA~\cite{akira_jgraphqa_2025} provided by JAMMEval~\cite{sugiura2026jammeval}, rather than the original JGraphQA benchmark. It evaluates Japanese chart and table understanding using images extracted from Japanese investor relations (IR) materials. The dataset comprises four types of figures: pie charts, line charts, bar charts, and tables.
\item
\textbf{JDocQA-Refined}, the refined version of JDocQA~\cite{onami2024jdocqa} provided by JAMMEval, rather than the original JDocQA benchmark. It is a Japanese document-reading VQA benchmark constructed from document images released by Japanese public institutions. Since the images are derived from PDFs, they tend to be high resolution and often require models to read small text within the image.
\end{itemize}

We report the arithmetic mean of these three benchmarks as \texttt{VQA-Overall}.

\subsection{Parsing-centric SFT}\label{sec:exp-parsing-centric-sft}

We first evaluate the effect of training on structured document parsing data alone. This parsing-centric SFT setting substantially improves DocParse-Overall from 70.43 to 86.56, showing that structured parsing capability can be effectively injected into the base model. However, this comes with clear forgetting: VQA-Overall drops from 0.863 to 0.826.

\begin{table}[h]
    \centering
    \caption{Comparison between the base model and the parsing-centric SFT model.}
    \label{tab:structured-sft-results}
    \fontsize{6}{6}\selectfont
    \setlength{\tabcolsep}{3pt}
    \renewcommand{\arraystretch}{1.2}
    \resizebox{\textwidth}{!}{%
        \begin{tabular}{l|cccc|cccc}
            \toprule
            \textbf{Model}
            & \textbf{Text\textsuperscript{Edit}$\downarrow$}
            & \textbf{Table\textsuperscript{TEDS}$\uparrow$}
            & \textbf{Formula\textsuperscript{CDM}$\uparrow$}
            & \makecell{\textbf{DocParse-}\\\textbf{Overall} $\uparrow$}
            & \makecell{\textbf{JA-Business-Doc-}\\\textbf{RQ-Bench} $\uparrow$}
            & \makecell{\textbf{JGraphQA-}\\\textbf{Refined} $\uparrow$}
            & \makecell{\textbf{JDocQA-}\\\textbf{Refined} $\uparrow$}
            & \makecell{\textbf{VQA-}\\\textbf{Overall} $\uparrow$} \\
            \midrule
            Base model
            & 0.3474 & 71.71 & 74.23 & 70.43
            & 0.895 & 0.883 & 0.810 & 0.863 \\
            Parsing-centric SFT
            & 0.1479 & 84.95 & 89.51 & 86.56
            & 0.834 & 0.872 & 0.770 & 0.826 \\
            \bottomrule
        \end{tabular}%
    }
\end{table}

We further compare thinking-mode and instruct-mode training using the same structured document parsing data. In thinking mode, the SFT target contains an English \texttt{<think>} trace followed by the final \texttt{document\_structure} JSON. In instruct mode, the reasoning trace is removed and the model is trained to generate the JSON directly. The training data source and optimization settings are otherwise kept the same.

Instruct-mode SFT achieves a slightly higher DocParse-Overall score, improving from 86.56 to 87.58. However, it is associated with substantial VQA degradation: VQA-Overall drops to 0.663, and JA-Business-Doc-RQ-Bench drops sharply from 0.895 for the base model to 0.384. This degradation is particularly important because JA-Business-Doc-RQ-Bench is designed to evaluate multi-step reasoning over visually rich Japanese business documents. The large drop suggests that removing the reasoning traces may disrupt behavior important for multi-step document VQA. We do not claim that this comparison fully isolates reasoning traces from all prompt- or template-related effects; rather, it shows that the direct-JSON instruct-mode setting is much less compatible with preserving VQA behavior in this experiment.

\begin{table}[h]
    \centering
    \fontsize{6}{6}\selectfont
    \caption{Comparison between thinking-mode and instruct-mode parsing-centric SFT.}
    \label{tab:structured-sft-thinking-instruct}
    \setlength{\tabcolsep}{3pt}
    \renewcommand{\arraystretch}{1.2}
    \resizebox{\textwidth}{!}{%
        \begin{tabular}{l|cccc|cccc}
            \toprule
            \textbf{Model}
            & \textbf{Text\textsuperscript{Edit}$\downarrow$}
            & \textbf{Table\textsuperscript{TEDS}$\uparrow$}
            & \textbf{Formula\textsuperscript{CDM}$\uparrow$}
            & \makecell{\textbf{DocParse-}\\\textbf{Overall} $\uparrow$}
            & \makecell{\textbf{JA-Business-Doc-}\\\textbf{RQ-Bench} $\uparrow$}
            & \makecell{\textbf{JGraphQA-}\\\textbf{Refined} $\uparrow$}
            & \makecell{\textbf{JDocQA-}\\\textbf{Refined} $\uparrow$}
            & \makecell{\textbf{VQA-}\\\textbf{Overall} $\uparrow$} \\
            \midrule
            Base model
            & 0.3474 & 71.71 & 74.23 & 70.43
            & 0.895 & 0.883 & 0.810 & 0.863 \\
            Parsing-centric SFT (thinking mode)
            & 0.1479 & 84.95 & 89.51 & 86.56
            & 0.834 & 0.872 & 0.770 & 0.826 \\
            Parsing-centric SFT (instruct mode)
            & 0.1677 & 85.74 & 93.77 & 87.58
            & 0.384 & 0.852 & 0.751 & 0.663 \\
            \bottomrule
        \end{tabular}%
    }
\end{table}

This result shows that direct JSON generation is effective for the parsing benchmark, but the instruct-mode setting is associated with substantially worse VQA performance, especially on multi-step document reasoning. This suggests that preserving English reasoning traces may help retain behavior important for document VQA. Therefore, the following experiments use thinking mode with English reasoning traces.

\subsection{Mixed SFT}\label{sec:exp-mixed-sft}

Mixed SFT is introduced to mitigate the forgetting observed in parsing-centric SFT. As shown in Table~\ref{tab:mixed-sft-results}, Mixed SFT achieves almost the same DocParse-Overall score as parsing-centric SFT, while substantially recovering VQA performance. DocParse-Overall remains essentially unchanged, moving from 86.56 to 86.60, while VQA-Overall improves from 0.826 to 0.844.

\begin{table}[h]
    \centering
    \caption{Comparison between parsing-centric SFT and mixed SFT.}
    \label{tab:mixed-sft-results}
    \fontsize{6}{6}\selectfont
    \setlength{\tabcolsep}{3pt}
    \renewcommand{\arraystretch}{1.2}
    \resizebox{\textwidth}{!}{%
        \begin{tabular}{l|cccc|cccc}
            \toprule
            \textbf{Model}
            & \textbf{Text\textsuperscript{Edit}$\downarrow$}
            & \textbf{Table\textsuperscript{TEDS}$\uparrow$}
            & \textbf{Formula\textsuperscript{CDM}$\uparrow$}
            & \makecell{\textbf{DocParse-}\\\textbf{Overall} $\uparrow$}
            & \makecell{\textbf{JA-Business-Doc-}\\\textbf{RQ-Bench} $\uparrow$}
            & \makecell{\textbf{JGraphQA-}\\\textbf{Refined} $\uparrow$}
            & \makecell{\textbf{JDocQA-}\\\textbf{Refined} $\uparrow$}
            & \makecell{\textbf{VQA-}\\\textbf{Overall} $\uparrow$} \\
            \midrule
            Base model
            & 0.3474 & 71.71 & 74.23 & 70.43
            & 0.895 & 0.883 & 0.810 & 0.863 \\
            Parsing-centric SFT
            & 0.1479 & 84.95 & 89.51 & 86.56
            & 0.834 & 0.872 & 0.770 & 0.826 \\
            Mixed SFT
            & 0.1718 & 87.50 & 89.48 & 86.60
            & 0.838 & 0.903 & 0.791 & 0.844 \\
            \bottomrule
        \end{tabular}%
    }
\end{table}

The key insight is that the structured document parsing/VQA trade-off is not fixed. By mixing VQA examples back into SFT, the model can retain most of the structured parsing gain while recovering much of the VQA capability lost under parsing-centric SFT. This makes Mixed SFT a better initialization for the subsequent RL stage.

\subsection{Parsing-centric RL}\label{exp-parsing-centric-rl}

We next evaluate whether parsing-centric RL can improve structured document parsing beyond the Mixed SFT checkpoint. The results show a clear difference between GSPO and DAPO.

GSPO does not exceed the Mixed SFT checkpoint in this setting, and DocParse-Overall decreases from 86.60 to 84.01. This does not mean that GSPO is inherently unsuitable for MoE models. Rather, in this particular task, the combination of sequence-level updates and KL anchoring appears to be too conservative to improve fine-grained parsing behavior. The task requires local improvements in JSON schema validity, element matching, bounding-box quality, table structure, and formula content after very long reasoning traces.

DAPO, in contrast, improves DocParse-Overall from 86.60 to 87.67 and Formula CDM from 89.48 to 92.87, achieving the best structured document parsing performance in this study. Under our training configurations, DAPO achieves a better parsing result than GSPO. Because the two configurations differ in several respects, including importance-ratio granularity, clipping, and KL regularization, this comparison does not isolate the contribution of any single design choice. The cost is a moderate VQA drop: VQA-Overall decreases from 0.844 to 0.825, reflecting the expected trade-off between moving beyond the SFT ceiling and preserving previously retained VQA capability.

\begin{table}[h]
    \centering
    \caption{Comparison of mixed SFT and parsing-centric RL methods.}
    \label{tab:rl-results}
    \fontsize{6}{6}\selectfont
    \setlength{\tabcolsep}{3pt}
    \renewcommand{\arraystretch}{1.2}
    \resizebox{\textwidth}{!}{%
        \begin{tabular}{l|cccc|cccc}
            \toprule
            \textbf{Model}
            & \textbf{Text\textsuperscript{Edit}$\downarrow$}
            & \textbf{Table\textsuperscript{TEDS}$\uparrow$}
            & \textbf{Formula\textsuperscript{CDM}$\uparrow$}
            & \makecell{\textbf{DocParse-}\\\textbf{Overall} $\uparrow$}
            & \makecell{\textbf{JA-Business-Doc-}\\\textbf{RQ-Bench} $\uparrow$}
            & \makecell{\textbf{JGraphQA-}\\\textbf{Refined} $\uparrow$}
            & \makecell{\textbf{JDocQA-}\\\textbf{Refined} $\uparrow$}
            & \makecell{\textbf{VQA-}\\\textbf{Overall} $\uparrow$} \\
            \midrule
            Base model
            & 0.3474 & 71.71 & 74.23 & 70.43
            & 0.895 & 0.883 & 0.810 & 0.863 \\
            Mixed SFT
            & 0.1718 & 87.50 & 89.48 & 86.60
            & 0.838 & 0.903 & 0.791 & 0.844 \\
            Mixed SFT + Parsing-centric RL (GSPO)
            & 0.1923 & 85.91 & 85.34 & 84.01
            & 0.808 & 0.852 & 0.778 & 0.813 \\
            Mixed SFT + Parsing-centric RL (DAPO)
            & 0.1712 & 87.25 & 92.87 & 87.67
            & 0.808 & 0.883 & 0.784 & 0.825 \\
            \bottomrule
        \end{tabular}%
    }
\end{table}

\subsection{Comparison with Open-Weight Models}\label{sec:comparision-with-open-weight-models}

Overall, the experiments indicate that Mixed SFT is effective for controlling forgetting, while DAPO-based parsing-centric RL is effective for pushing structured parsing performance beyond the SFT ceiling. The final model therefore uses the Mixed SFT + DAPO-based parsing-centric RL recipe. As the final outcome of this study, we release the mixed SFT + DAPO-based parsing-centric RL model as Stockmark-Nemotron-3-Nano-Omni-JapanDocReader. This model was selected because it achieved the strongest structured document parsing performance among the variants studied in this work, while still retaining part of the original document VQA capability of the base model. We compare the released model with several open-weight models on OmniDocBench-JASyn in Table~\ref{tab:open-model-comparison}.

\begin{table}[h]
\centering
\caption{Structured document parsing performance comparison with open-weight models on OmniDocBench-JASyn.}
\label{tab:open-model-comparison}
\fontsize{6}{6}\selectfont
\setlength{\tabcolsep}{3pt}
\renewcommand{\arraystretch}{1.2}
\resizebox{\textwidth}{!}{%
    \begin{tabular}{l|c|cccc}
        \toprule
        \textbf{Model}
        & \makecell{\textbf{DocParse-}\\\textbf{Overall} $\uparrow$}
        & \textbf{Text\textsuperscript{Edit}$\downarrow$}
        & \textbf{Table\textsuperscript{TEDS}$\uparrow$}
        & \textbf{Formula\textsuperscript{CDM}$\uparrow$}
        & \textbf{Reading Order\textsuperscript{Edit}$\downarrow$} \\
        \midrule
        Qwen3.6-27B~\cite{qwen36}
        & 84.32
        & 0.1247
        & 86.23
        & 79.19 
        & 0.2548 \\

        Qwen3.5-27B~\cite{qwen35}
        & 83.85
        & \textbf{0.1154}
        & 80.81
        & 82.27 
        & 0.2599 \\
        
        gemma-4-31B-it~\cite{gemmateam2026gemma4}
        & 81.42
        & 0.1943
        & 83.10
        & 80.60 
        & 0.2598 \\
        
        Nemotron-3-Nano-Omni-30B-A3B-Reasoning~\cite{deshmukh2026nemotron}
        & 70.43
        & 0.3474
        & 71.71
        & 74.23 
        & 0.4157 \\
        \midrule
        Stockmark-Nemotron-3-Nano-Omni-JapanDocReader
        & \textbf{87.67}
        & 0.1712
        & \textbf{87.25}
        & \textbf{92.87}
        & \textbf{0.2366} \\
        \bottomrule
    \end{tabular}%
}

\end{table}

Stockmark-Nemotron-3-Nano-Omni-JapanDocReader achieves a DocParse-Overall score of 87.67, outperforming the other open-weight models in this comparison. The improvement is especially clear on Formula CDM, where the released model reaches 92.87, indicating substantially stronger structured parsing capability for Japanese documents containing mathematical expressions. It also obtains the best Table TEDS score, 87.25, suggesting that the model has learned to produce more accurate table structures in the target JSON format. In addition to the DocParse-Overall components, Stockmark-Nemotron-3-Nano-Omni-JapanDocReader also achieves the best Reading Order Edit score, 0.2366. Although reading order is not a component of DocParse-Overall, this result is consistent with the hypothesis that bbox-based element matching provides a cleaner learning signal than index-wise matching. During RL, content and localization sub-scores are computed after same-class bbox-based element matching, rather than by directly comparing elements at the same array index. This matching step reduces artificial order noise caused by a single insertion or deletion in the \texttt{document\_structure} array. However, this observation is correlational: a controlled ablation is required to isolate the effect of the matching strategy on Reading Order Edit. Qwen3.5-27B and Qwen3.6-27B achieve lower Text Edit scores, indicating stronger pure text extraction performance. However, structured document parsing requires more than text extraction alone: the model must jointly produce valid JSON, layout elements, bounding boxes, table HTML, formula LaTeX, and picture descriptions. From this broader perspective, Stockmark-Nemotron-3-Nano-Omni-JapanDocReader achieves the strongest overall structured parsing performance among the compared open-weight models.

This result demonstrates that mixed SFT followed by DAPO-based parsing-centric RL can substantially improve the Japanese structured document parsing capability of Nemotron-3-Nano-Omni. Compared with the original Nemotron-3-Nano-Omni-30B-A3B-Reasoning model, DocParse-Overall improves from 70.43 to 87.67, and Reading Order Edit improves from 0.4157 to 0.2366. Importantly, this improvement is achieved in a setting where we explicitly consider the trade-off between capability injection and forgetting: the goal is not only to teach the model to output structured JSON, but also to preserve as much of its original document VQA capability as possible.

\section{Conclusion}\label{sec:conclusion}

We introduced Stockmark-Nemotron-3-Nano-Omni-JapanDocReader, a post-trained Nemotron-3-Nano-Omni model for Japanese document understanding. The model is obtained by mixed SFT followed by DAPO-based parsing-centric RL, and is designed for structured document parsing through capability injection and forgetting control. It learns to produce page-level \texttt{document\_structure} JSON outputs, including layout elements, bounding boxes, tables, formulas, picture descriptions, and reading order, while preserving Japanese document VQA behavior as much as possible.

The experiments show that parsing-centric SFT is an effective way to inject structured parsing capability into the base model. It substantially improves DocParse-Overall and teaches the model the target JSON schema and page-level parsing behavior, but it also causes measurable VQA forgetting. The comparison between thinking-mode and instruct-mode SFT further shows that removing English reasoning traces is associated with substantial degradation on multi-step VQA, especially on document reasoning questions that require comparison, aggregation, conditional reasoning, and calculation. Mixed SFT substantially reduces this forgetting while maintaining almost the same structured document parsing performance as parsing-centric SFT. This indicates that capability injection and VQA preservation are not strictly conflicting objectives: by mixing Japanese document VQA data back into SFT, the model can retain much of its original reasoning behavior while acquiring structured parsing ability.

Finally, DAPO-based parsing-centric RL further improves structured document parsing beyond the mixed SFT ceiling and produces the final released model. This improvement comes with additional VQA drift, highlighting the central trade-off studied in this report: stronger parsing capability injection can move the model away from its original VQA distribution. The results also show that parsing-centric RL depends heavily on reward design and prompt filtering. A matched, application-aware reward, together with variance-based filtering for continuous structured rewards, is important for making RL effective in long-reasoning structured document parsing tasks.

\section*{Acknowledgements}\label{sec:acknowledgements}

In this experiment, we used 8×B300 Blackwell Ultra compute resources provided by NVIDIA, and conducted training on NVIDIA Brev. We sincerely thank everyone at NVIDIA for supporting the large-scale SFT and RL experiments.

\bibliography{main}

\clearpage 
\newpage
\appendix

\section*{Appendix}
\label{sec:appendix}

\section{Structured Document Parsing Prompt}
\label{sec:prompt}

For structured document parsing, we use a fixed Japanese prompt across evaluation and model comparison. The prompt asks the model to extract the document structure from the input image and to return the result in a JSON format. The expected output consists of a top-level \texttt{document\_structure} array, where each element contains a class label, a bounding box, contents, and an optional caption for picture elements. The fixed prompt is shown below.

\begin{Shaded}
\begin{Highlighting}
\NormalTok{画像に含まれるドキュメントの構造をJSON形式で抽出してください。}
\NormalTok{出力フォーマット:}
\NormalTok{\{}
\NormalTok{  "document\_structure": [}
\NormalTok{    \{}
\NormalTok{      "class": "title" | "heading" | "text" | "table" | "list" | "picture" | "formula",}
\NormalTok{      "bbox": [x1, y1, x2, y2],}
\NormalTok{      "contents": "内容（pictureの場合は画像内容の説明）",}
\NormalTok{      "caption": "pictureのキャプション文字（オプション）"}
\NormalTok{    \}}
\NormalTok{  ]}
\NormalTok{\}}
\NormalTok{classの種類: title（タイトル）、heading（見出し）、text（本文）、table（表）、list（リスト）、}
\NormalTok{picture（画像）、formula(数式)}
\NormalTok{bboxは左上(x1,y1)と右下(x2,y2)の座標です。}
\end{Highlighting}
\end{Shaded}

The class set contains seven document element types: \texttt{title}, \texttt{heading}, \texttt{text}, \texttt{table}, \texttt{list}, \texttt{picture}, and \texttt{formula}. Bounding boxes are represented as (\texttt{[x\_1, y\_1, x\_2, y\_2]}), corresponding to the top-left and bottom-right coordinates of each element. For \texttt{picture} elements, the \texttt{contents} field is used to describe the visual content, while \texttt{caption} stores the caption text when it is present in the document.

\section{Qualitative Examples}
\label{sec:qualitative-behavior}

\subsection{Structured Document Parsing}\label{sec:structured-parsing-samples}

\begin{figure}[H]
\centering
\includegraphics[width=0.98\linewidth]{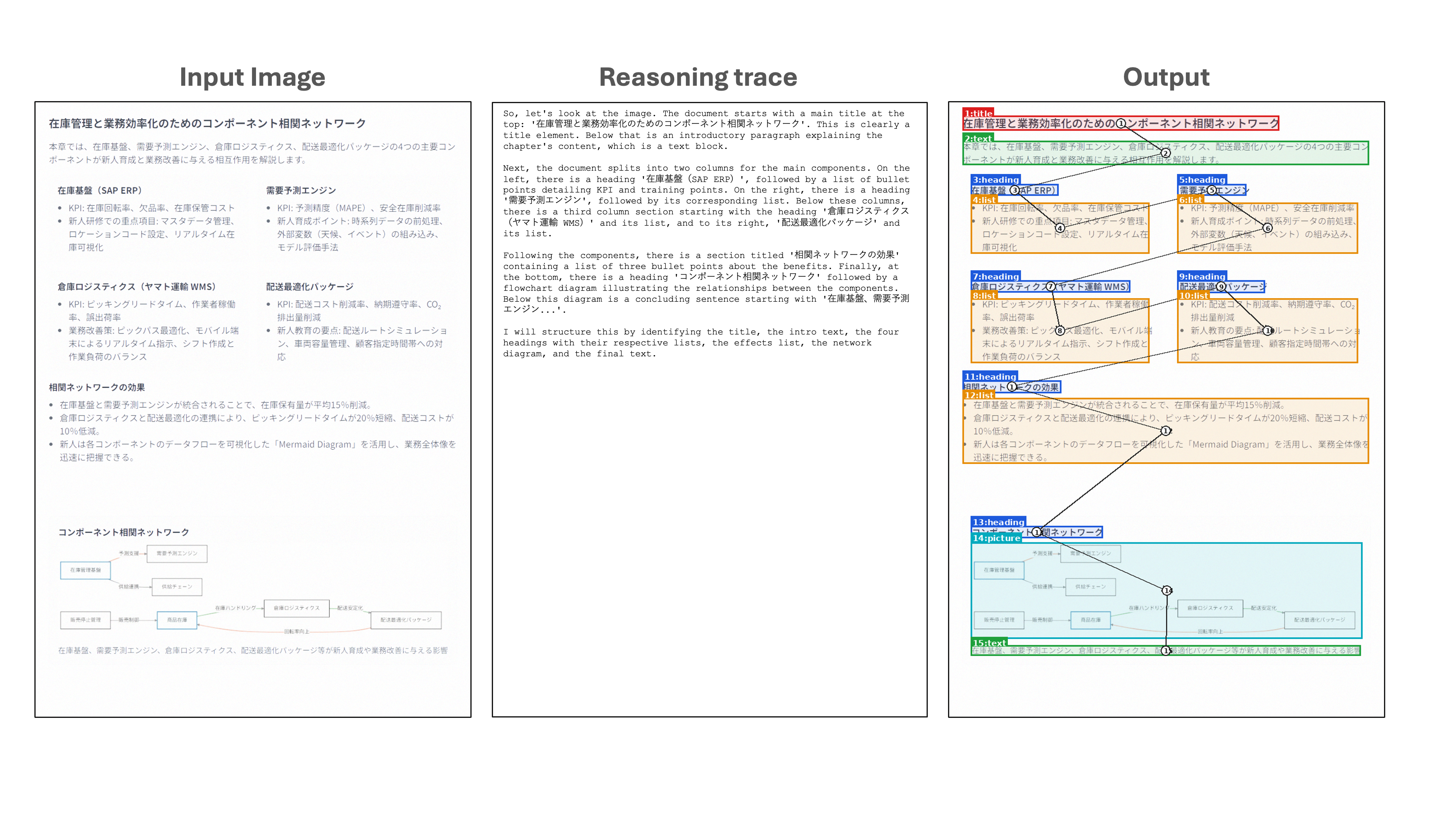}
\end{figure}

\begin{figure}[H]
\centering
\includegraphics[width=0.98\linewidth]{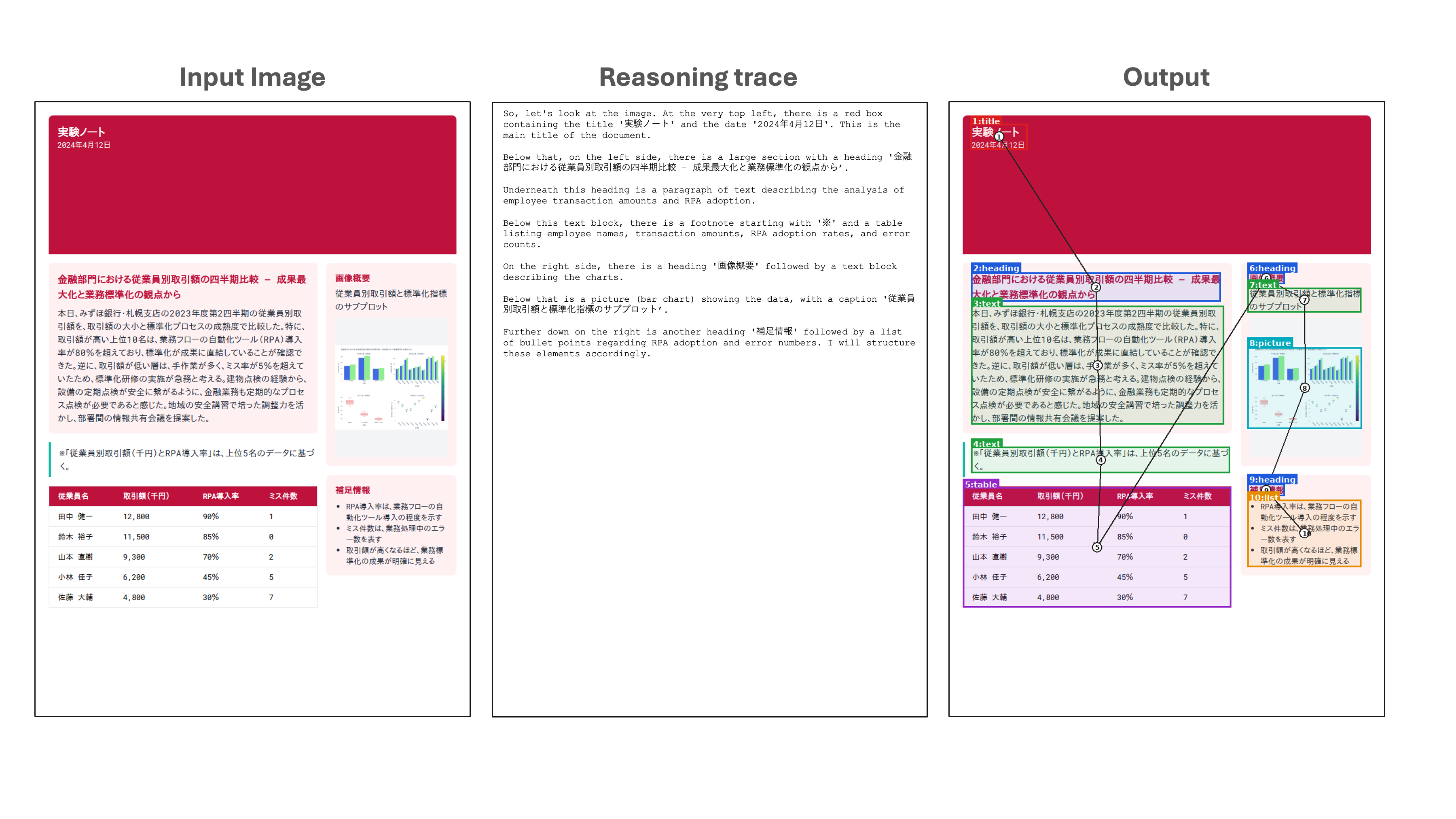}
\end{figure}

\begin{figure}[H]
\centering
\includegraphics[width=0.98\linewidth]{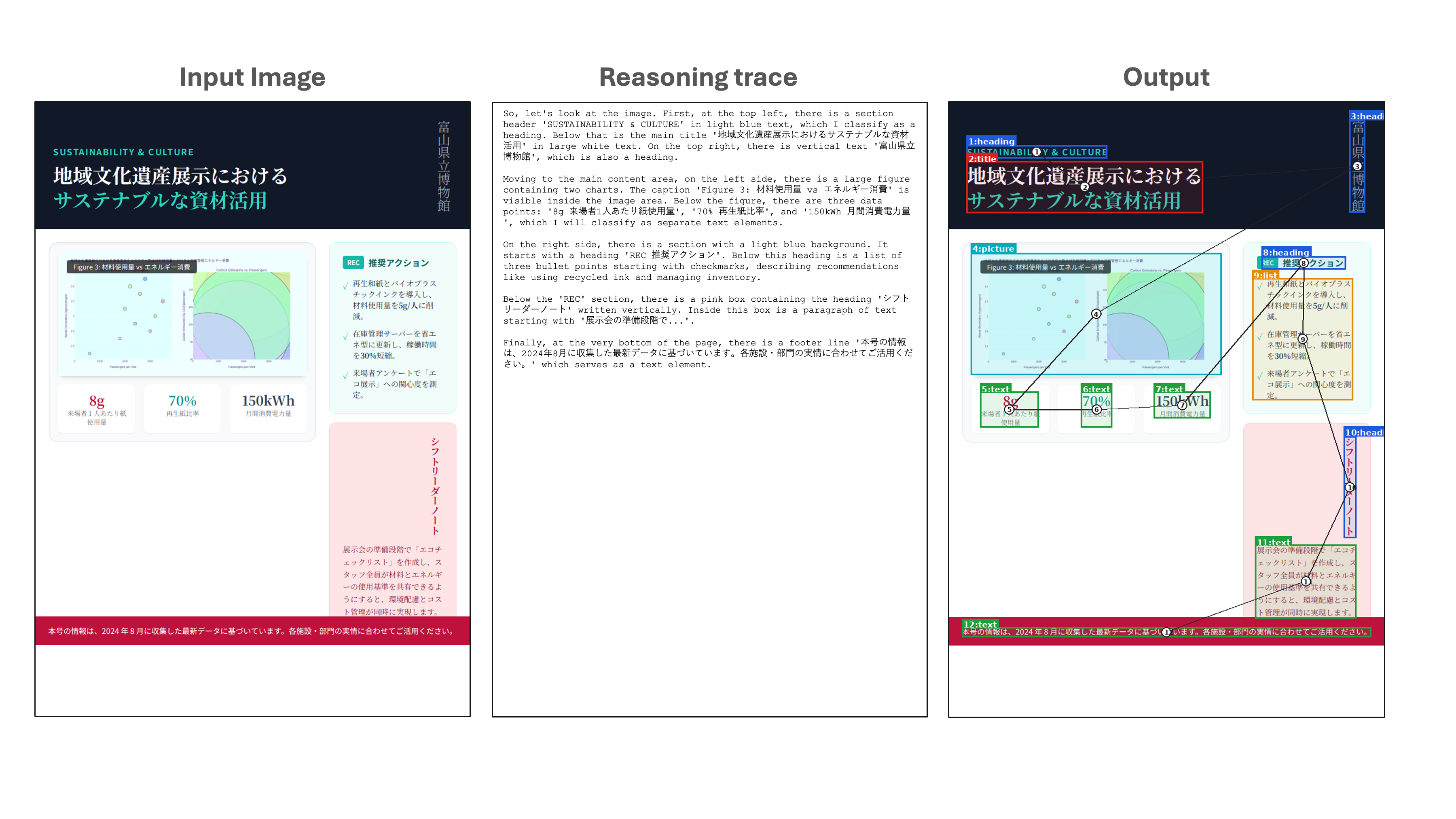}
\end{figure}

\begin{figure}[H]
\centering
\includegraphics[width=0.98\linewidth]{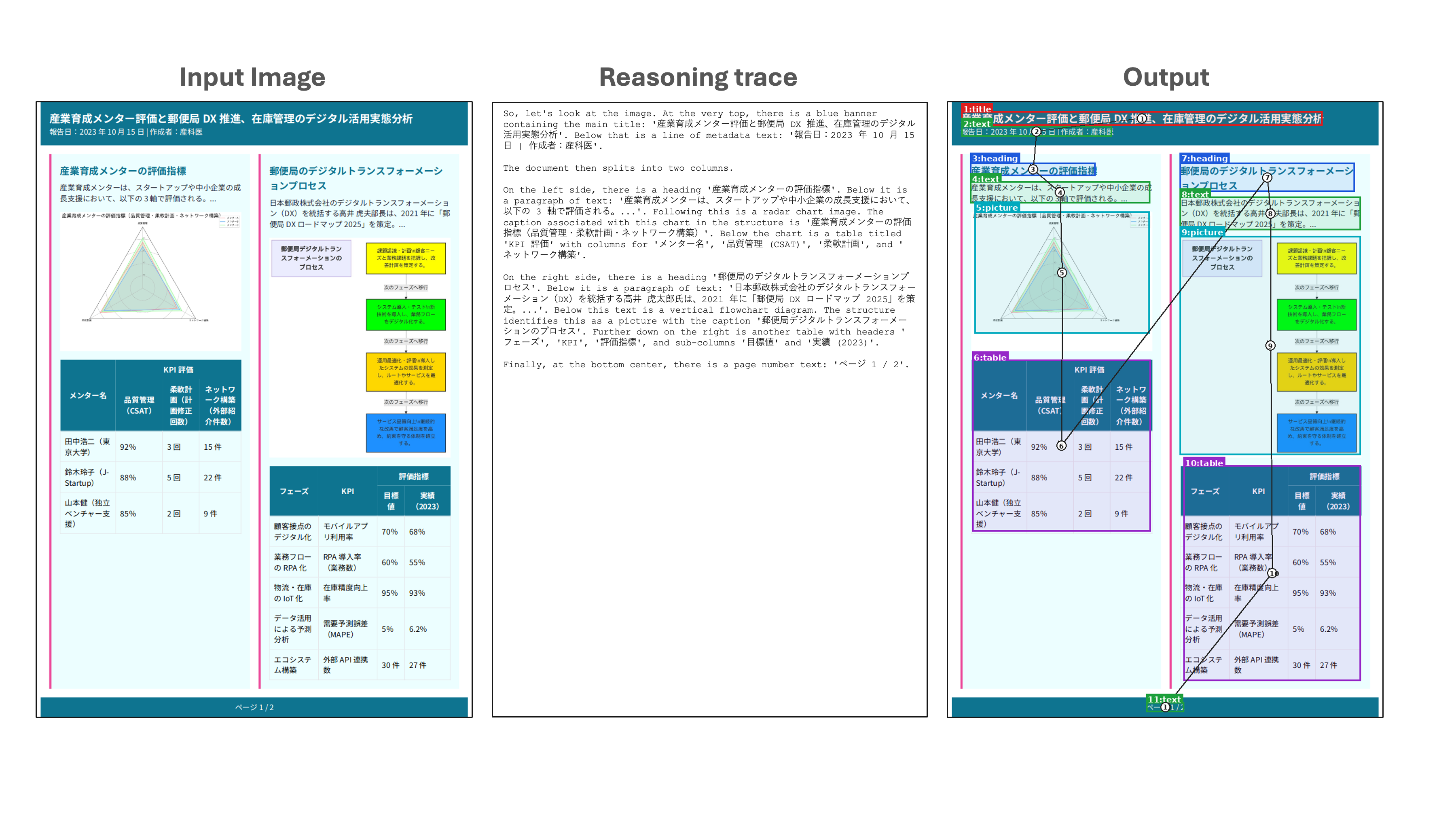}
\end{figure}

\begin{figure}[H]
\centering
\includegraphics[width=0.98\linewidth]{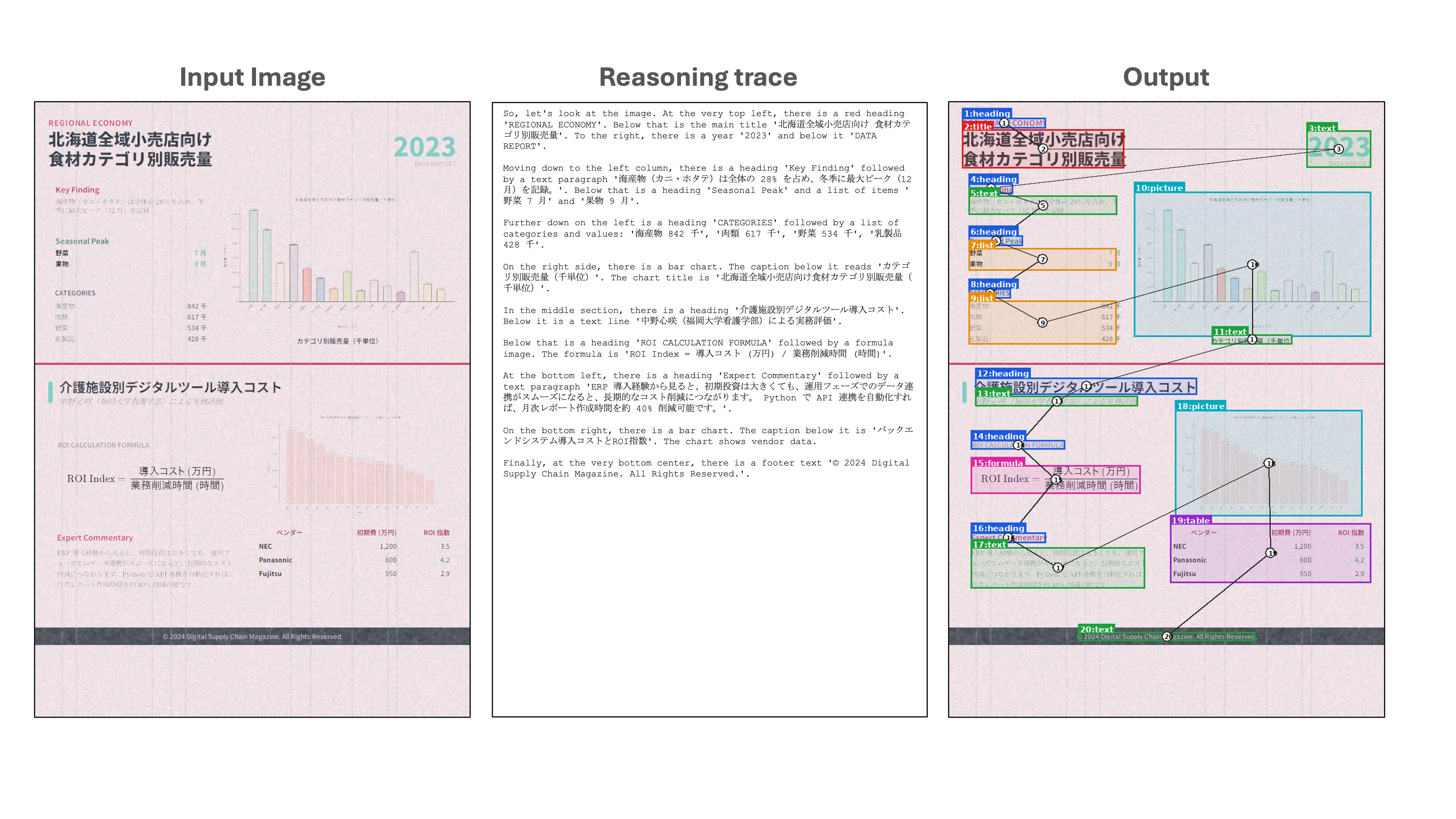}
\end{figure}

\begin{figure}[H]
\centering
\includegraphics[width=0.98\linewidth]{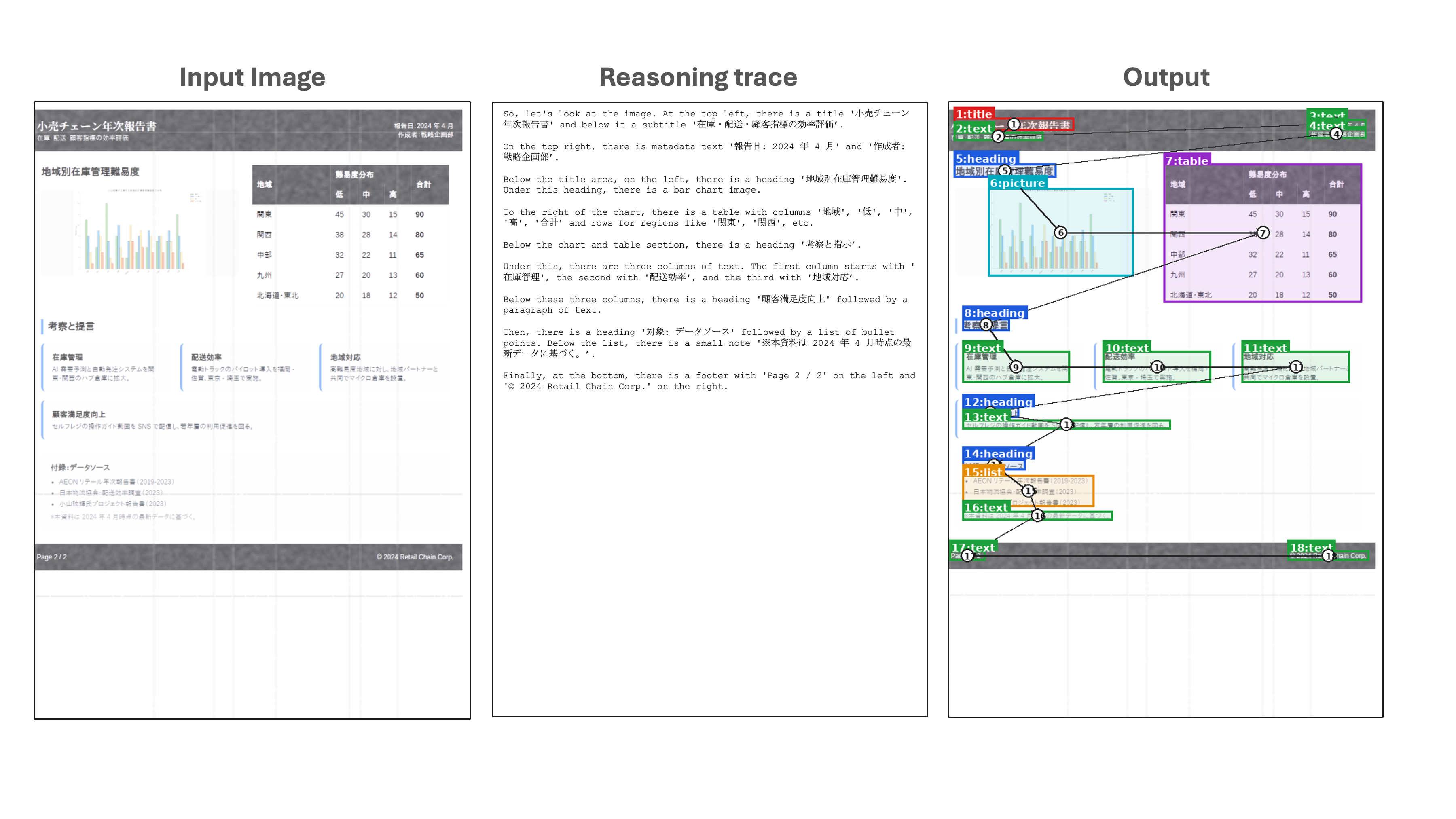}
\end{figure}

\begin{figure}[H]
\centering
\includegraphics[width=0.98\linewidth]{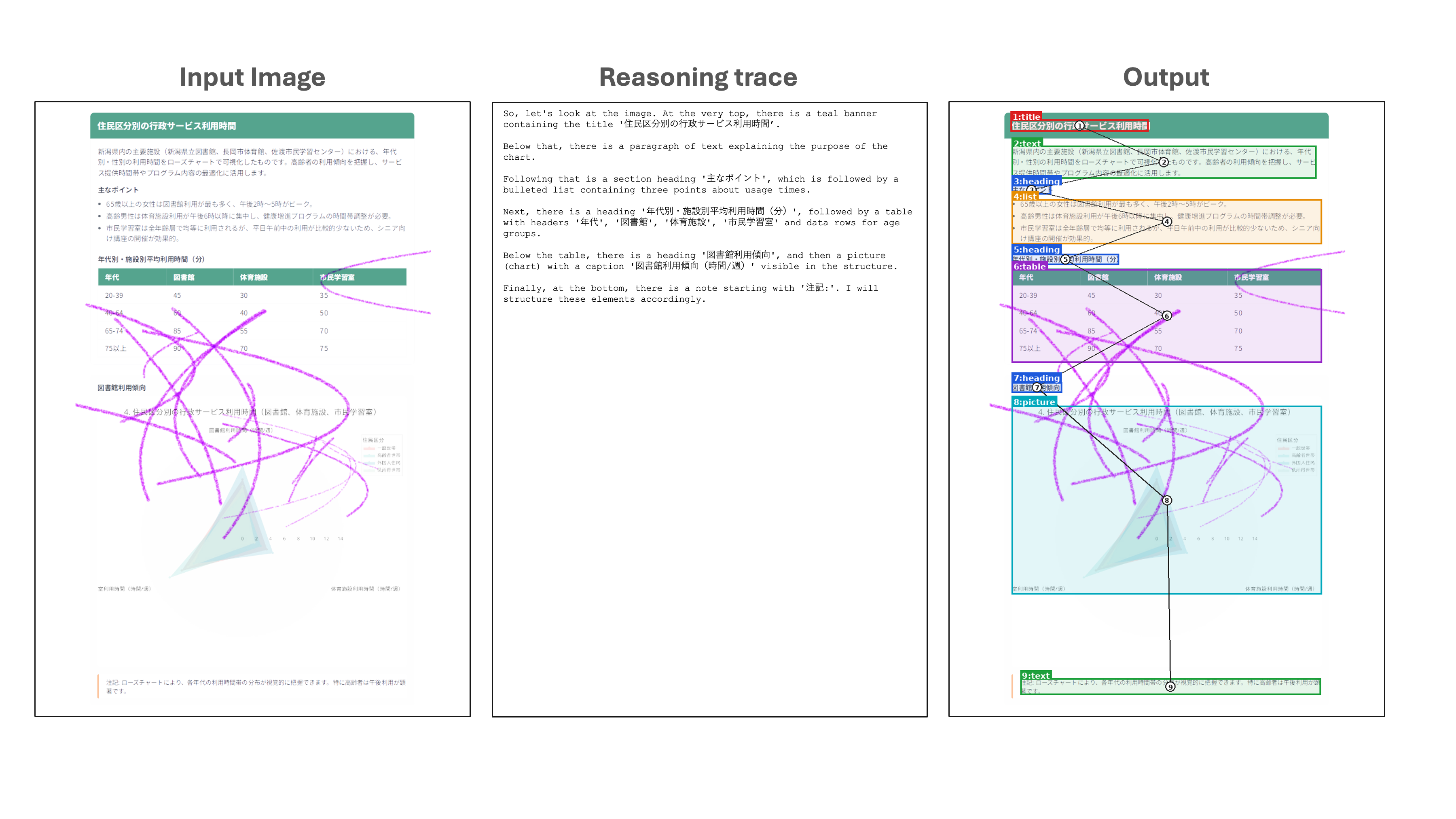}
\end{figure}

\subsection{Japanese Document VQA}\label{sec:japanese-document-vqa-samples}

\begin{figure}[H]
\centering
\includegraphics[width=0.98\linewidth]{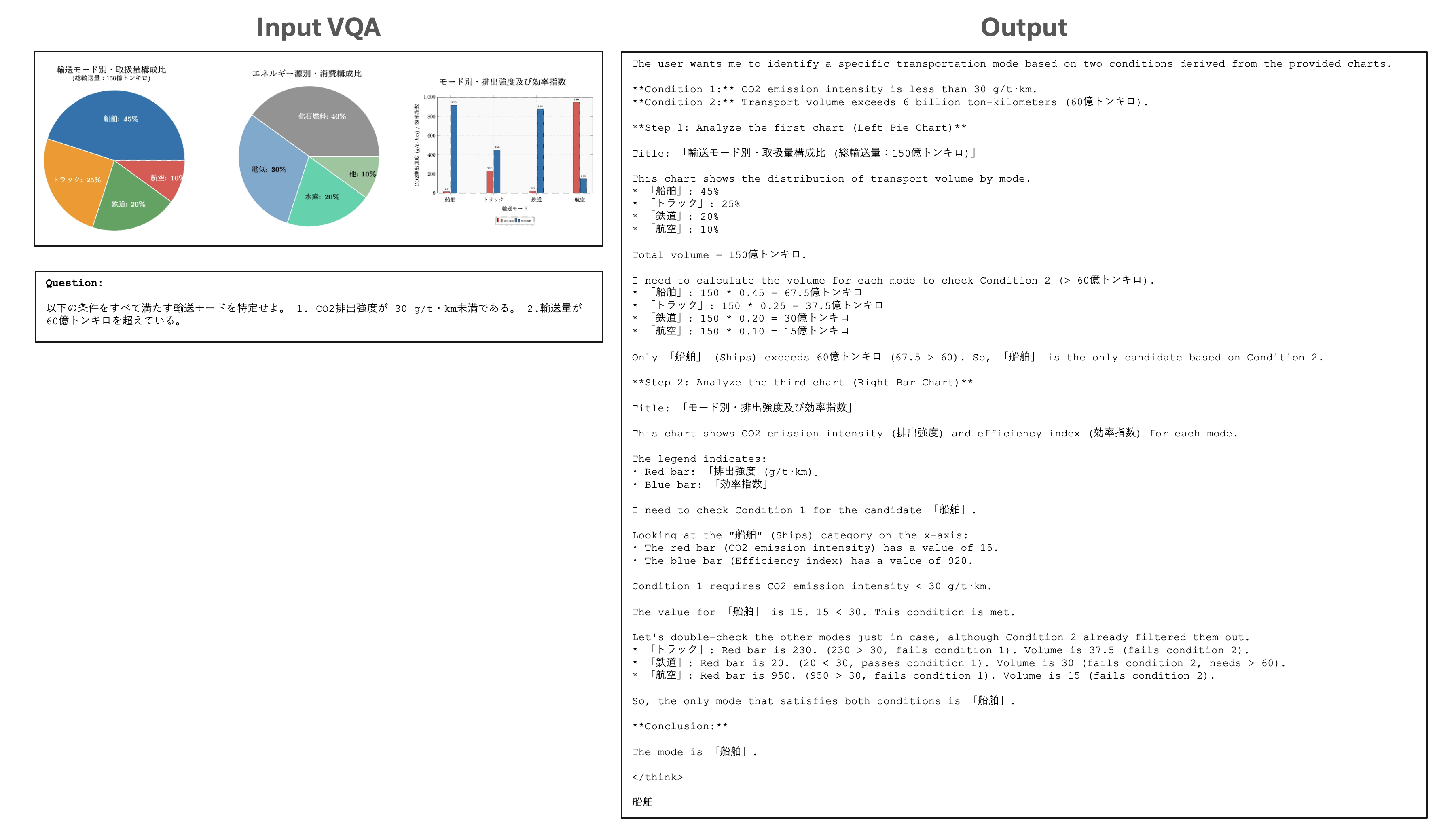}
\end{figure}

\begin{figure}[H]
\centering
\includegraphics[width=0.98\linewidth]{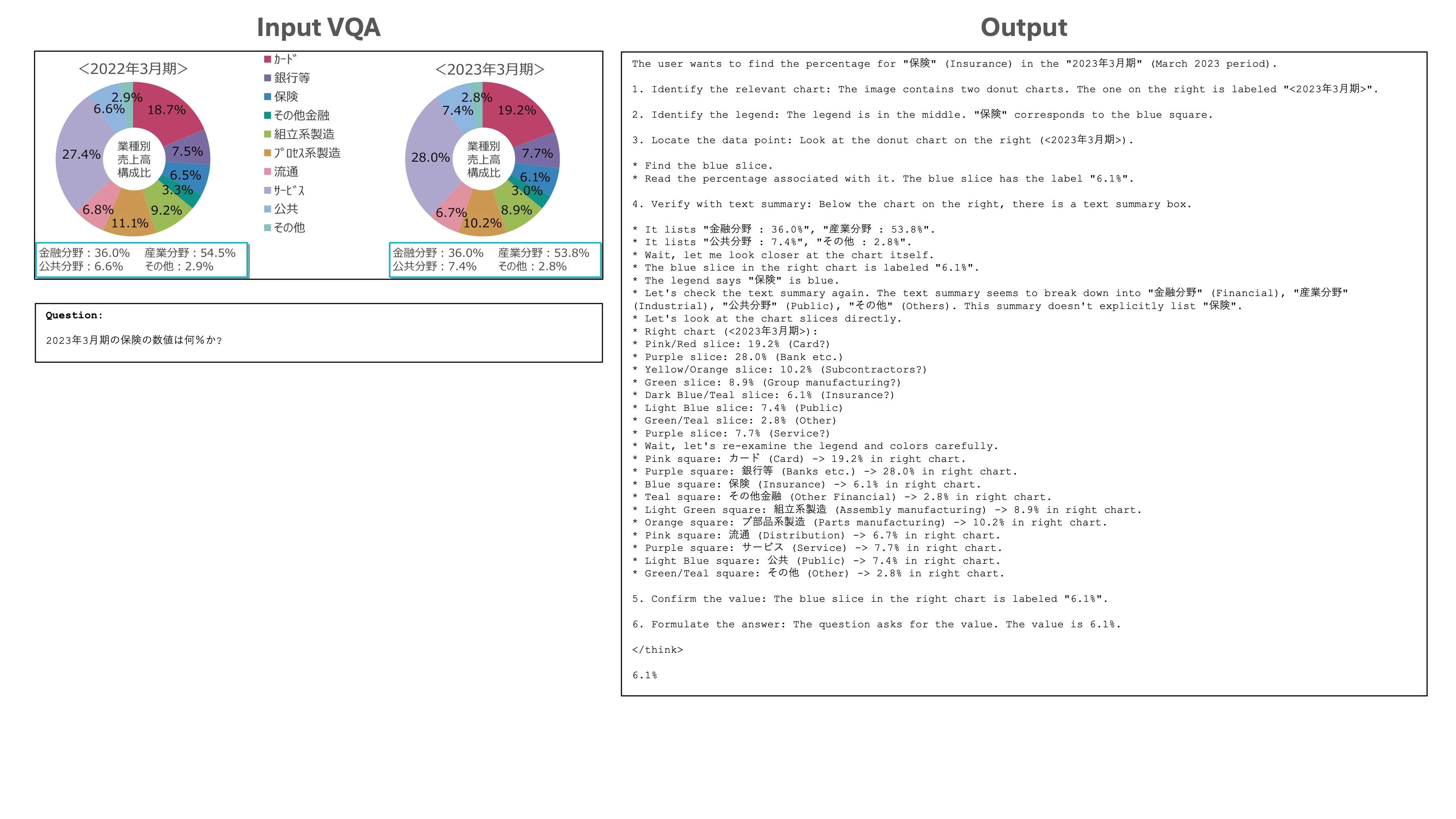}
\end{figure}

\begin{figure}[H]
\centering
\includegraphics[width=0.98\linewidth]{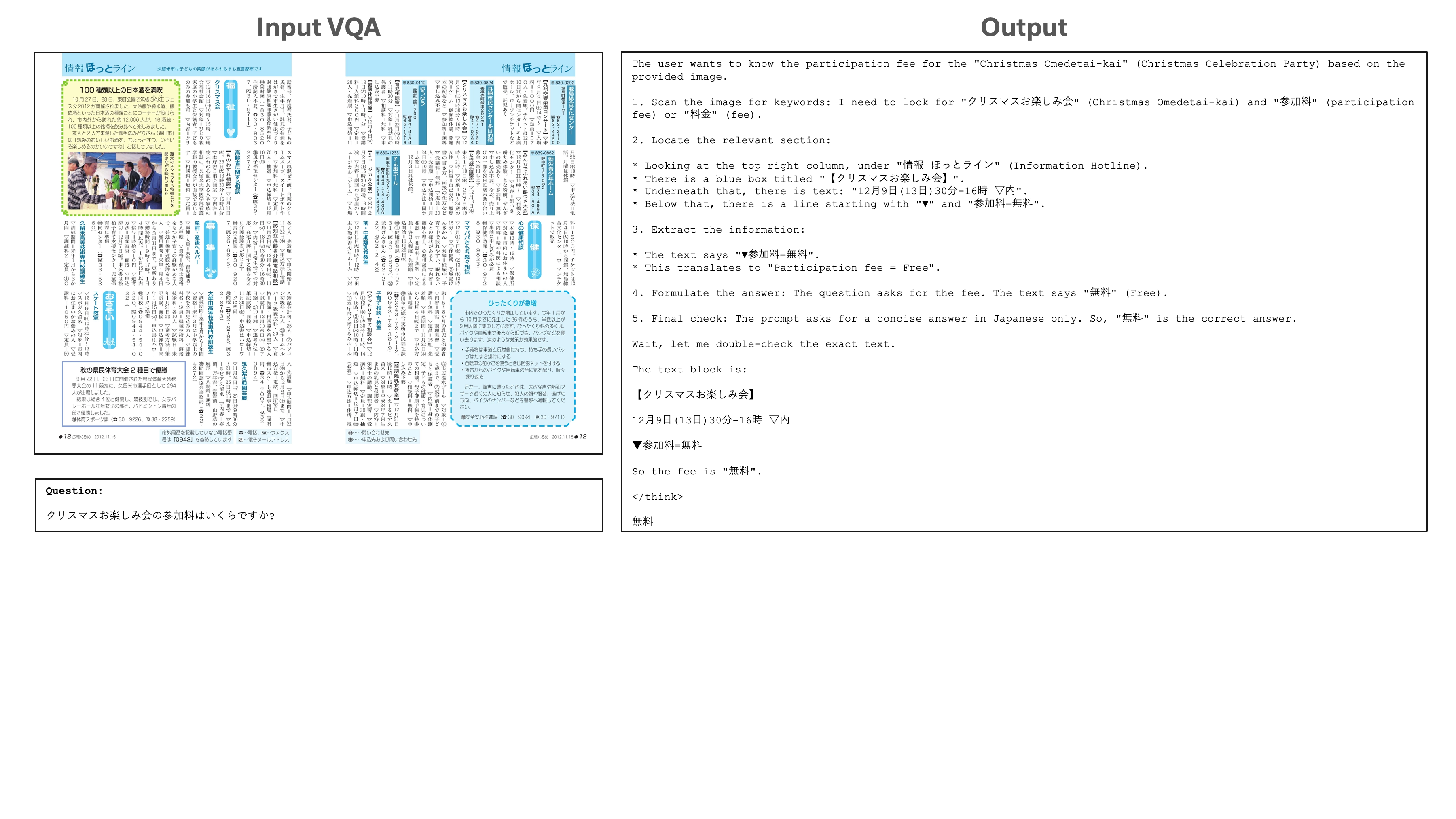}
\end{figure}

\end{document}